\documentclass[11pt]{article}

\usepackage[final]{acl}

\usepackage{times}
\usepackage{latexsym}

\usepackage[T1]{fontenc}

\usepackage[utf8]{inputenc}

\usepackage{microtype}

\usepackage{inconsolata}

\usepackage{graphicx}

\usepackage{rotating}
\usepackage{soul}
\usepackage{tikz, makecell, colortbl}
\usetikzlibrary{arrows,decorations.markings,shapes,matrix,positioning, calc, fit,decorations.pathreplacing,angles,quotes,backgrounds}
\usepackage{siunitx}
\usepackage{multirow}
\usepackage{subcaption}
\usepackage{enumitem}
\usepackage{framed}
\usepackage{longtable } %
\usepackage{xspace}
\usepackage{pdflscape}
\usepackage{array}
\usepackage{etoolbox}
\usepackage{tablefootnote}
\usepackage{threeparttable}
\usepackage{xurl}

\usepackage{hyperref}

\usepackage{float}
\usepackage{csquotes}
\usepackage{placeins}
\usepackage{booktabs}

\renewcommand{\sectionautorefname}{\S\kern-0.2em}
\renewcommand{\subsectionautorefname}{\S\kern-0.2em}
\renewcommand{\subsubsectionautorefname}{\S\kern-0.2em}

\newcolumntype{P}[1]{>{\raggedright\arraybackslash}p{#1}}
\newcolumntype{Q}[1]{>{\centering\arraybackslash}p{#1}}
\newcolumntype{R}[1]{>{\raggedleft\arraybackslash}p{#1}}

\definecolor{customrowcolor}{gray}{0.9}
\RequirePackage{color}\definecolor{RED}{rgb}{1,0,0}\definecolor{BLUE}{rgb}{0,0,1}

\RequirePackage[normalem]{ulem}
\providecommand{\DIFadd}[1]{{\protect\color{blue}\uwave{#1}}} %
\providecommand{\DIFdel}[1]{{\protect\color{red}\sout{#1}}} %

\usepackage[most]{tcolorbox}

\newtcolorbox[list inside=prompt,auto counter,number within=section]{prompt}[1][]{
    colbacktitle=black!60,
    fonttitle=\small,
    coltitle=white,
    fontupper=\footnotesize,
    boxsep=4pt,
    left=0pt,
    right=0pt,
    top=0pt,
    bottom=0pt,
    boxrule=1pt,
    float,
    floatplacement=tb,
    #1,
}

\makeatletter
\providecommand*{\tcb@cnt@promptautorefname}{Prompt}
\makeatother

\newcommand{\modelgenerated}{LLM-generated\xspace}

\newcommand{\Description}[1]{}

\title{Can You Make It Sound Like You? Post-Editing LLM-Generated Text for Personal Style}

\author{
  Connor Baumler \quad
  Calvin Bao \quad
  Huy Nghiem \quad
  Xinchen Yang \\
  {\bf Marine Carpuat} \quad
  {\bf Hal Daum\'e III} \quad\\
  University of Maryland \\
  \texttt{baumler@cs.umd.edu} \quad \texttt{me@hal3.name} %
}

\begin{document}
\maketitle
\begin{abstract}
Despite the growing use of large language models (LLMs) for writing tasks, users may hesitate to rely on LLMs when personal style is important. 
Post-editing LLM-generated drafts or translations is a common collaborative writing strategy, but it remains unclear whether users can effectively reshape LLM-generated text to reflect their personal style. 
We conduct a pre-registered online study ($n=81$) in which participants post-edit LLM-generated drafts for writing tasks where personal style matters to them.
Using embedding-based style similarity metrics, we find that post-editing increases stylistic similarity to participants' unassisted writing and reduces similarity to fully LLM-generated output. 
However, post-edited text still remains stylistically closer in style to LLM text than to participants' unassisted control text, and it exhibits reduced stylistic diversity compared to unassisted human text. 
We find a gap between perceived stylistic authenticity and model-measured stylistic similarity, with post-edited text often perceived as representative of participants' personal style despite remaining detectable LLM stylistic traces.
\end{abstract}

\section{Introduction}

Today, many people use large language models (LLMs) to assist with routine workplace writing and other professional communication.~\cite{LIANG2025101366}
However, their use in personal writing tasks---where authenticity and faithfulness to the author's personal style is especially salient---remains controversial.
For example, online discussions about AI-assisted wedding vows show mixed reactions: while some view such use as ``icky'' or as signaling insufficient care or effort, others argue that using AI to produce an initial draft that is then personalized is no different from consulting example vows and adapting them.\footnote{For example, see \href{https://www.reddit.com/r/wedding/comments/11td56w/would_it_bother_you_if_your_partner_used_ai_to/}{this Reddit discussion on AI-assisted wedding vows}.}

These intuitions align with prior findings that people's acceptance of AI assistance depends on the writing task and where in the writing process assistance is applied. \citet{reza2025cowritingaihumanterms} show that in tasks where personal style is important, writers are more receptive to AI support in planning and organizing ideas, but resist assistance in translating ideas into language or revising text, as these stages directly shape stylistic expression.

Motivated by these findings, we ask whether writers can accept AI involvement in initial drafting if they can post-edit the draft to reintroduce their personal style.
In a pre-registered\footnote{\url{https://aspredicted.org/ck6xk4.pdf}} and IRB-approved\footnote{University of Maryland IRB package number 2072441-4} online study ($n=81$), participants completed writing tasks in which personal style matters to them, such as wedding vows or an apology letter. In treatment tasks, participants provided planned content details and then post-edited an \modelgenerated draft.\footnote{The collected data is available at \url{https://github.com/ctbaumler/personal_style_postedit}.}

Using embedding-based style similarity metrics, which have been shown to substantially capture linguistic style--often to a higher degree than human assessments~\cite{rexha_authorship_2018}---we find that participants can post-edit \modelgenerated drafts to make them sound significantly more like their own unassisted writing and significantly less like \modelgenerated text. Participant self-reports align with this pattern: they perceive that their post-edited text is more stylistically similar to themselves and is more usable for the writing task. Participants also expressed a preference for using this workflow for future personal writing with the most prominent obstacle to adoption being the lack of originality of the \modelgenerated text. 

Despite these stylistic improvements and participant satisfaction, post-edited drafts remained substantially more similar stylistically to \modelgenerated text than to participants' unassisted writing and were overall less stylistically diverse than unassisted human writing. 
Nevertheless, participants perceived their post-edited drafts as equally representative of their personal style and suitable for the writing task. Even when embedding-based metrics detect residual LLM stylistic cues, participants considered the resulting text authentic and usable. These results highlight a disconnect between model-measured stylistic similarity and user-perceived authenticity, emphasizing that participants can achieve outcomes they find acceptable without fully aligning \modelgenerated drafts with their unassisted writing.

\section{Background}\label{sec:rlw}

\subsection{Defining and Measuring Style} %

\paragraph{What is style?} 

\citet{crystal_investigating} define writing style as ``a selection of language habits, the occasional linguistic idiosyncrasies which characterize an individual's uniqueness''. \citet{Biber_Conrad_2009} treat style as patterns of lexical and grammatical variation distinct from register (variation driven by situational context) and genre (variation driven by text format). We adopt Crystal's view of style as individual linguistic habits, focusing on personal writing tasks rather than cross-domain comparisons.

Stylistic choices are socially meaningful: writers adjust style to signal identity or group affiliation \cite{Khalid2020StyleMI} and to manage audience expectations and engagement~\cite{10.1145/3706598.3714034, 10.1145/3613904.3642697}. These considerations raise the risk that AI assistance in co-writing may implicitly push writers toward a more generic or homogenized style that conflicts with their intended audience or personal style.

\paragraph{How is style measured?}

Computational approaches to measuring style have taken multiple forms. Early work in stylometry captures author- or community-specific fingerprints using features such as function-word distributions, part-of-speech patterns, and character- or word-level statistics~\cite{argamon_gender_2003, herrman_2015, Khalid2020StyleMI}. Another approach, common in style transfer literature, treats style as a discrete attribute, such as formality, simplicity, or sentiment, and evaluates generated text using classifiers for the target attribute~\cite{xu-2017-shakespeare, tikhonov-etal-2019-style, jin-etal-2022-deep}. An alternative method learns continuous embeddings from authorship data using objectives that pull together texts expected to share style (e.g., written by the same author) and push apart texts that differ in style, even when content varies~\cite{rivera-soto-etal-2021-learning, wegmann-etal-2022-author, patel-etal-2025-styledistance}.

\paragraph{What is the style of \modelgenerated text?}
A growing body of work has documented systematic stylistic differences between \modelgenerated and human-written text~\citep[e.g.][]{dentella2025chatgptgeneratedtextsauthorshiptraits, shaib-etal-2024-detection, richburg-etal-2024-automatic}. These differences include lower-level patterns such as lexical choice and syntactic templates~\cite{gray2024chatgptcontaminationestimatingprevalence, shaib-etal-2024-detection}, as well as higher-level tendencies such as reduced specificity or reliance on generic phrasing~\cite{chakrabarty2025aiwritingsalvagedmitigating}. Such stylistic signatures can be detected both automatically and by expert human readers~\citep[e.g.][]{emi2024technicalreportpangramaigenerated, russell-etal-2025-people}.

Beyond detecting LLM-associated stylistic traces, recent work has examined whether \modelgenerated text can be ``salvaged'' into higher-quality writing through editing. \citet{chakrabarty2025aiwritingsalvagedmitigating} show that expert human edits can mitigate issues such as clich\'ed phrasing, lack of detail, and awkward constructions, producing writing that is preferred by creative writing experts over the model's default output. While this work focuses on aligning AI writing with expert norms of quality, we ask whether post-editing can shift \modelgenerated text toward an authentic representation of an individual writer's personal style.

\subsection{Co-Writing With LLMs}

\citet{FlowerHayes1981} propose a cognitive process theory of writing in which writers iteratively engage in \textit{planning} what to express, \textit{translating} those plans into natural language, \textit{reviewing} the text, and \textit{monitoring} when to switch between processes. Building on this framework, prior work has examined how LLMs may be introduced at different stages of the writing process through various interaction strategies~\cite{reza2025cowritingaihumanterms}. One widely studied method for human-AI co-creation of text is \textit{post-editing}, in which humans revise machine-generated drafts. This approach has a long history in machine translation research as a means of improving output quality and speed~\cite{knight1994automated, simard-etal-2007-statistical}.

Prior work shows that co-writing with LLMs can influence various aspects of human writing. These effects include changes in stylistic features such as lexical choice \cite{padmakumar2024does}, as well as content-related characteristics including expressed opinions \cite{Jakesch2023CoWriting}, stereotypes \cite{10.1145/3772318.3790733}, and self-presentation \cite{10.1145/3544549.3585893}. Together, this literature suggests that LLMs can shape both \textit{what} is written and \textit{how} it is expressed.

\citet{reza2025cowritingaihumanterms} find that writer preferences for AI assistance depend on both the goals of the writing task and the stage of the writing process at which AI intervenes. In writing scenarios where content is the primary contribution, writers tend to resist AI assistance during planning but are more receptive to AI support for translating ideas into language. In contrast, in writing scenarios where form is more important and that heavily involve one's ``unique voice and style'', writers are willing to use AI during planning, but resist AI involvement in translating or reviewing text. Similarly, \citet{10.1145/3711020} find that AI writing assistance is least harmful to perceived authenticity when used during early, exploratory stages of idea development rather than in producing the final draft.

Together, this work suggests that writers are generally receptive to AI support during early stages of writing but resist AI involvement in later stages when stylistic control is most important. We build on this observation by examining whether post-editing \modelgenerated drafts allows writers to express their personal style, and how they perceive the value of this workflow in settings where stylistic authenticity matters.

\section{Research Hypotheses}\label{sec:rqs}

Our study focuses on participants' ability and desire to express their unique writing style. They write or post-edit text in scenarios in which they perceive writing with their authentic style to be important. AI systems have distinctive, often unhumanlike writing styles, which are unlikely to match a user's style without intervention.

To this end, we formulate two main preregistered hypotheses.\footnote{These hypotheses have been lightly edited from preregistration for clarity.} First, we consider whether participants can post-edit \modelgenerated text to sound authentically like themselves in writing scenarios where personal style is important to them.\footnote{Style encompasses multiple independent features, so, as we will later see, texts can become more similar to one reference (the participant's own control writing, LLM-generated text, etc) independently of change with respect to another.}  %

\begin{description}[nolistsep,noitemsep]
    \item[H1:]Participants' AI-assisted writing (through post-editing) matches their own natural, unassisted writing style.
    \begin{description}[nolistsep,noitemsep]
        \item[H1a:] Participants' AI-assisted writing sounds more like their unassisted writing after post-editing than before.
        \item[H1a$^\prime$:] Post-editing will make participants' AI-assisted writing sound more like their own unassisted writing than it does like other participants' unassisted writing.
        \item[H1b:] Participants' AI-assisted writing sounds less like \modelgenerated writing after post-editing than before.
        \item[H1c:] Participants' post-edited writing sounds less like \modelgenerated writing than their unassisted writing.
    \end{description}
\end{description}

Second, we investigate the style of the resulting post-edited text themselves and whether these texts are best categorized as a distinct and detectable ``third'' variety that is neither fully human-authored nor fully \modelgenerated.

\begin{description}[nolistsep,noitemsep]
    \item[H2:] Post-edited text has a distinct and consistent style.
    \begin{description}[nolistsep,noitemsep]
    \item[H2a:] Participants' LLM-assisted writing (through post-editing) has a more homogenous style than their non-AI assisted writing
    \item[H2b:] Participants' LLM-assisted writing has a less homogeneous style than \modelgenerated writing
    \item[H2c:] This post-edited style is distinct enough that after-post editing, participants will sound less like their own unassisted writing than they do like other participants' post-edited writing. 
    \end{description}
\end{description}

The hypotheses above focus on model-measured changes in style. We additionally consider whether these measurements match participants' perceptions of style.

\begin{description}[nolistsep,noitemsep]
    \item[H3:] Participants agree with automatic metrics at identifying whether they have successfully post-edited LLM-assisted writing to sound like themselves.
\end{description}

Beyond these hypotheses, we present further preregistered analyses of participants' perceptions of the usability of post-edited text and of potential inequities in post-editing effort across demographic groups as well as additional non-preregistered exploratory analyses in \autoref{sec:other_results}.

\section{Study Procedure}\label{sec:study_design}

\begin{figure*}[tb]
    \centering
    \includegraphics[width=\linewidth, trim={0 347 25 0},clip]{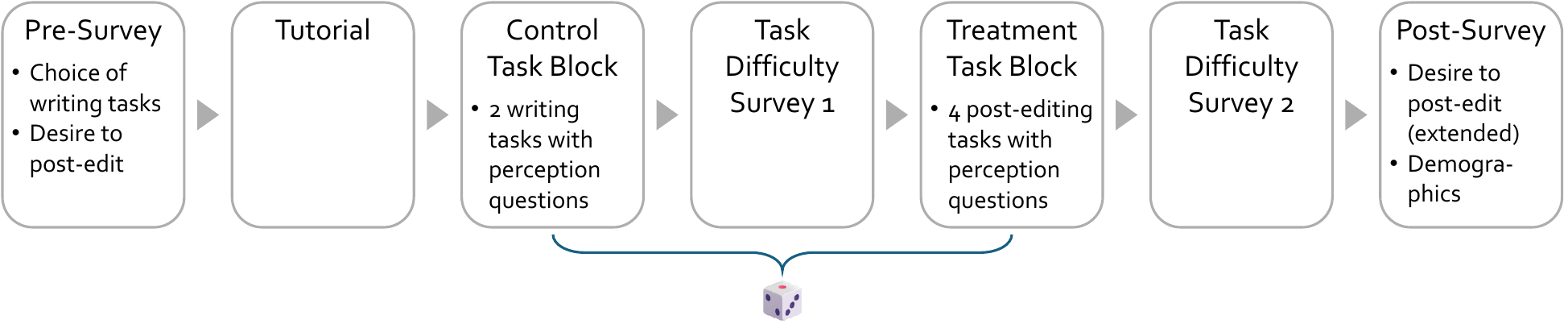}
    \caption{Study overview: pre-survey, tutorial, two randomized task blocks (treatment and control) with task difficulty surveys, and a post-survey. See \autoref{sec:screenshots} for interface screenshots.}
    \label{fig:study_phases}
    \Description{Summary of study structure: tutorial, pre-survey, writing tasks, task difficulty survey 1, writing tasks, task difficulty survey 2, and post-survey}
\end{figure*}

\begin{figure*}[tb]
    \centering
    \includegraphics[width=\linewidth, trim={0 111 25 0},clip]{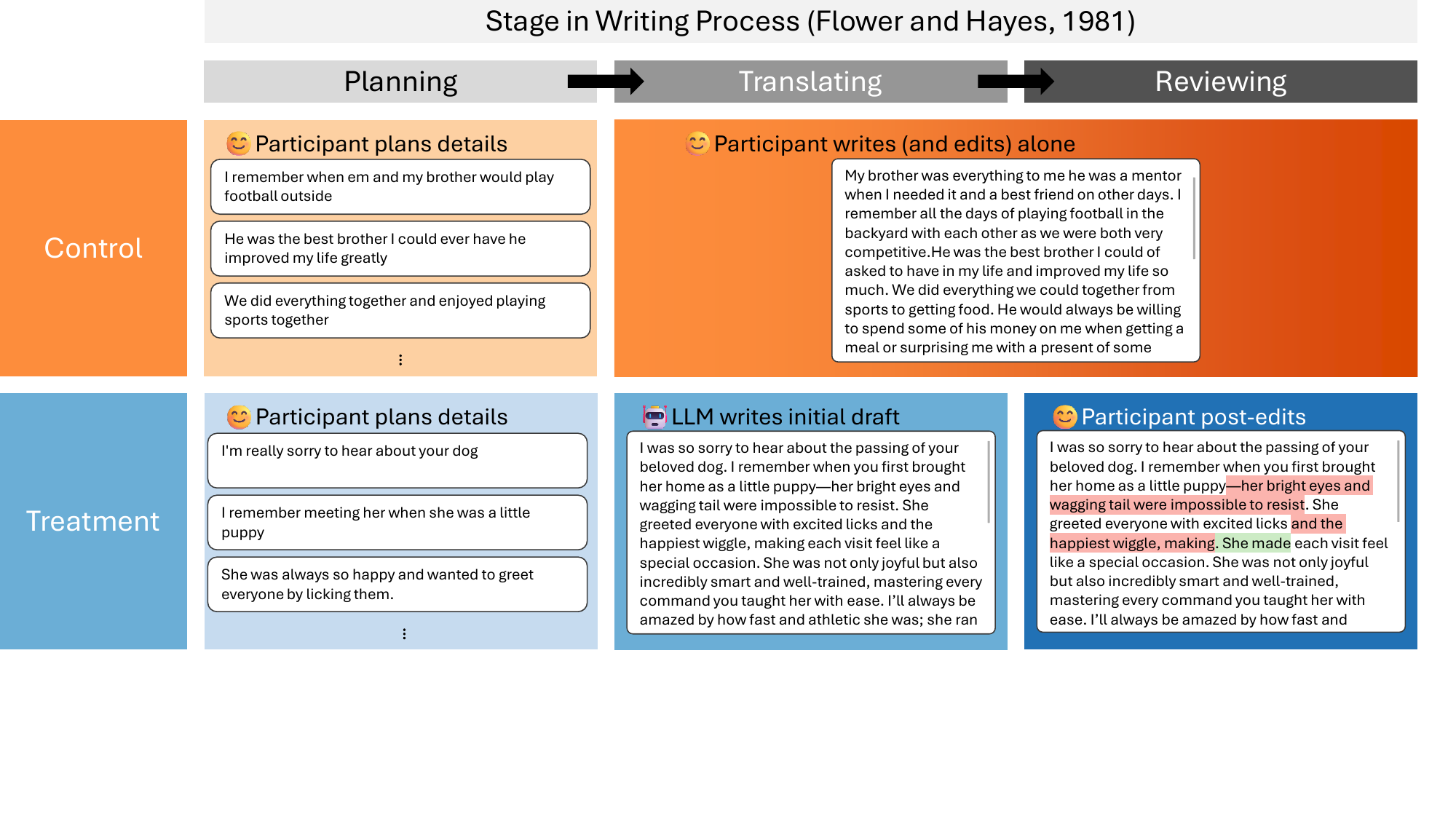}
    \caption{The main writing task. In both treatment and control blocks, participants plan details alone (top and bottom, left). In the treatment block, an LLM generates an initial draft (bottom, middle) which participants post-edit (bottom, right), and in the control block, participants write alone (top, middle and right). See \autoref{sec:screenshots} for a full set of interface screenshots and \autoref{tab:example_output} for example text from the treatment condition.} %
    \label{fig:writing_procedure}
\end{figure*}

As illustrated in \autoref{fig:study_phases}, participants completed pre- and post-surveys, task difficulty surveys, a brief tutorial, and two randomized writing task blocks (treatment and control). Using this procedure, we recruit $100$ participants of which we include $81$ in our final analysis (See \autoref{sec:participant_demo} for more details about participants and compensation.).

\paragraph{Pre- and Post-Surveys.} In the pre-survey, participants selected six of eight writing tasks\footnote{See \autoref{tab:tasks} for the task options and \autoref{sec:formative} for details on how the set of tasks was determined.} and rated their willingness to use AI for writing and ability to post-edit for style. These questions were repeated in the post-survey, which also included questions about participants' desire to post-edit \modelgenerated text in the future, factors influencing this decision (e.g., quality, privacy), etc. Full survey items are provided in \autoref{sec:screenshots}.

\paragraph{Task Difficulty Surveys.} After each task block, participants completed a short survey assessing task difficulty (\autoref{fig:mid_survey}) through modified NASA TLX questions~\cite{HART1988139}.

\paragraph{Tutorial.} Participants completed a brief interactive tutorial introducing the process of planning details and then writing or editing a draft.

\paragraph{Writing Task and Task-Level Survey.} For each selected writing task, participants provided at least $30$ characters of planned details, such as where the couple met to include in wedding vows (\autoref{fig:writing_procedure}, \autoref{fig:task}).\footnote{To preserve privacy, they are asked to provide fake details.} These details were used to generate the LLM drafts as we are interested in the edits participants make to fix \textit{stylistic issues} in the drafts, not to fix the \textit{factual content}. The six selected writing tasks were randomized between the treatment and control blocks. 

\newpage %

In the \textit{treatment} block, participants edited the \modelgenerated draft to reflect their personal style, spending at least two minutes\footnote{We did not set a minimum amount of edits, as participants may have felt that the \modelgenerated text already sufficiently matched their personal style.} before continuing. In the \textit{control} block, participants wrote at least $150$ characters directly from their plans without seeing any \modelgenerated text. Task-level questions collected prior experience and perceived importance of personal style for the given task. Participants also rated the stylistic authenticity of their writing on a $5$-point likert scale: in the control condition, this referred only to their own fully human-written text, and in the treatment condition, to both the original \modelgenerated draft and their post-edited version. In the treatment condition, participants additionally indicated the types of edits they made.

\begin{prompt}[title={Prompt \thetcbcounter: Prompt for wedding vows writing task}, label=prompt:eg_prompt]
Please help me write a draft wedding vows. Only respond with the draft vows with no additional content or explanation. Please do not include any placeholder text such as ``[name''. The final draft should be about 150 words. Here are some details to help you write:\\
\texttt{[Bulleted list of participant-written details]}
\end{prompt}

\section{Methods}

We use LLMs both to generate initial drafts in the treatment condition and to measure the style of treatment and control texts.

\subsection{Generation of LLM Drafts}\label{sec:llm_methods}
LLM drafts were generated using \textsc{GPT}-o4-mini~\cite{o4mini}. The model was prompted with the type of writing task, the participant-written list of details,\footnote{Under this design, it is possible that the LLM could pick up on stylistic cues in the list of details and tailor its output accordingly. We consider this concern in \autoref{sec:llm_consistency} and find no evidence of such stylistic personalization in our study.} and instructions including approximate word count (see \autoref{prompt:eg_prompt}).

\subsection{Measuring Style}

Our analyses largely use measures of stylistic similarity as outcomes. We estimate stylistic similarity between texts using the cosine similarity between their \texttt{LUAR} embeddings~\cite{rivera-soto-etal-2021-learning}. As we discuss in \autoref{sec:emb_selection}, we selected this style embedding model as it was the most performant on an authorship identification task on our control data. While it may not be the current top-performing model on standard authorship benchmarks, we find that it most effectively captures the stylistic differences among participants within the specific writing-task domains of this study.

We also analyze AI detector scores via \textsc{Pangram}~\cite{emi2024technicalreportpangramaigenerated}. These scores estimate the likelihood of a text being AI-authored, similar to assessing the stylistic similarities between a given text against an \modelgenerated text.

\section{Results}\label{sec:quant_results}

In this section, we examine how effectively participants post-edited \modelgenerated text to sound like themselves (\autoref{sec:h1}) and the style of post-edited text itself (\autoref{sec:h2}). We also discuss participants' perception of style (\autoref{sec:h3}), the  features participants do and do not post-edit (\autoref{sec:qual}), and participants' reasons to use or not use LLM drafts in future writing where personal style is important (\autoref{sec:why_postedit}).

Unless otherwise stated, group comparisons were performed using permutation tests (see \autoref{sec:permutation_tests}), with corresponding effect sizes reported as Hedges' $g$~\cite{hedges1981} with $95\%$ confidence intervals calculated from $1000$ bootstrap samples. Associations were analyzed using repeated-measures correlation~\cite{bakdash2017repeated}. Claims of statistical significance for preregistered analyses refer to results surviving Benjamini–Hochberg correction \cite{1995benjaminicontrolling}, controlling the false discovery rate at 
$q=0.05$. Raw $p$-values are reported throughout.\footnote{Exploratory analyses were not included in this multiple-testing correction and are identified as such.}

\subsection{Effectiveness of Post-Editing in Improving Stylistic Authenticity}\label{sec:h1}

\begin{figure}[tb]
    \centering
    \includegraphics[width=0.95\linewidth, clip,trim=5 5 5 5]{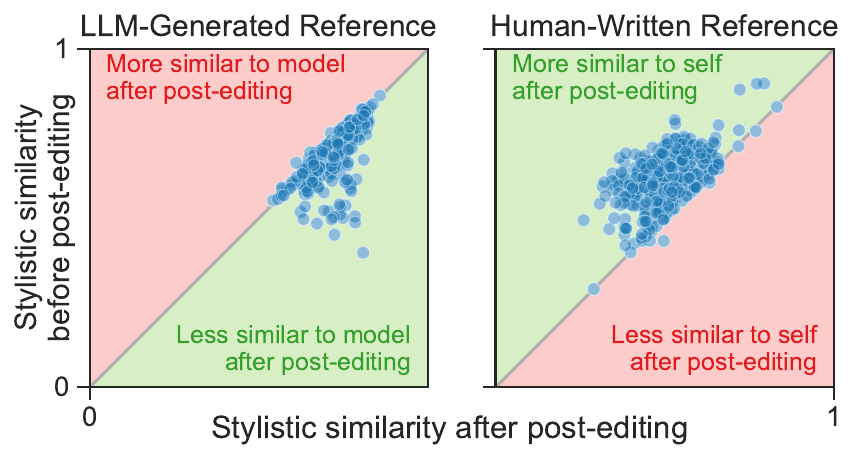}
    \caption{Similarity to \modelgenerated text (left) and control text (right) before and after post-editing. Through post-editing, text is more similar to control text (H1a) and less similar to \modelgenerated text (H1b).}
    \Description{}
    \label{fig:similarity_scatter}
\end{figure}

First, we consider how well participants were able to post-edit \modelgenerated text to match their personal writing style (H1).\footnote{This analysis includes all treatment observations. We explore in \autoref{sec:style_importance} how these effects may vary on tasks where personal style was somewhat less important to participants.} After post-editing, participants' treatment text is significantly more stylistically similar to their control writing ($p=.0002$, $g=0.55$, $95\%$ CI: [$0.38$, $0.71$], \autoref{fig:similarity_scatter} right) and significantly less similar to \modelgenerated text (H1b, $p=.0002$, $g=-0.41$, $95\%$ CI: [$-0.44$, $-0.39$], \autoref{fig:similarity_scatter} left). This shift away from stylistic similarity to \modelgenerated text is also observed using \textsc{Pangram} AI-detector scores (H1b, $p=.0002$, $g=-0.45$, $95\%$ CI: [$-0.55$, $-0.35$]). 

We also see that the stylistic shift through post-editing moves participants' style more toward their own unassisted style than it does to other participants' style (H1a$^\prime$, $p=.0002$, $g=-0.56$, $95\%$ CI: [$-0.7$, $-0.43$], \autoref{fig:similarity_scatter_other_pe} left). That is, the effects seen in H1a and H1b would not be well explained by saying that participants post-editing their text to sound generically more ``human''. In this case, we would have expected post-editing to lead to roughly equal increases in stylistic similarity regardless of which human author we compare to.

However, despite significant evidence that participants could post-edit \modelgenerated text to better match the style of their control writing, participants' post-edited text remained significantly more stylistically similar to \modelgenerated text than to their unassisted control text (H1c, $p=.0002$, $g=-1.43$, $95\%$ CI: [$-1.55$, $-1.32$], \autoref{fig:similarity_scatter_pre_post} right), implying that participants were not able to fully remove LLM stylistic cues from the drafts.

\subsection{Post-Editing and Stylistic Homogeneity}\label{sec:h2}

\begin{figure}[tb]
    \centering
    \includegraphics[width=0.8\linewidth, clip, trim=2 2 2 2]{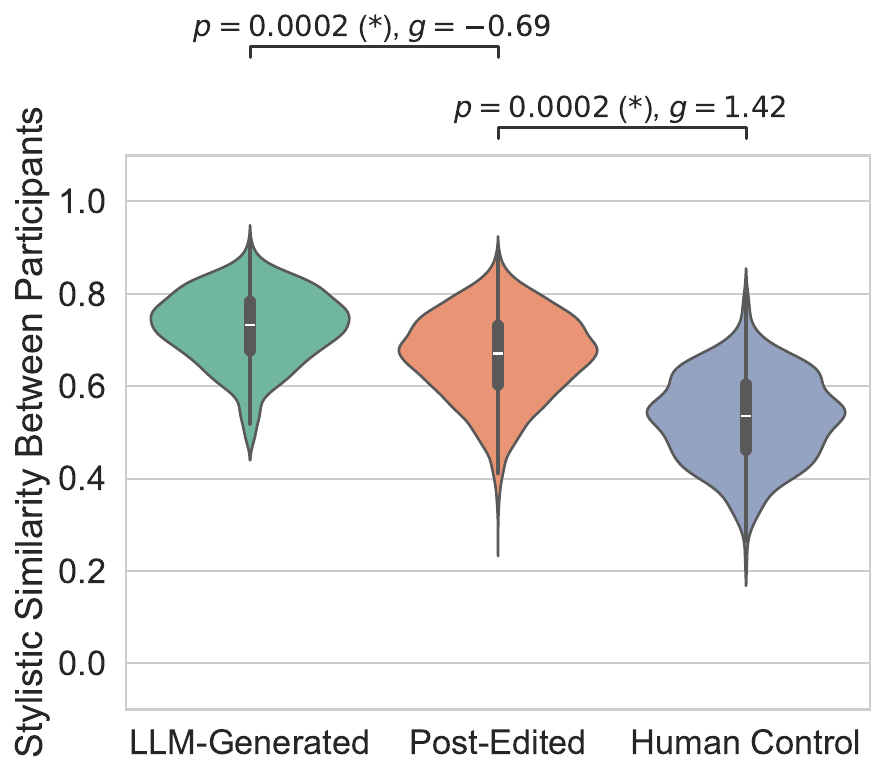}
    \caption{Stylistic similarity between pairs of text from \textit{different} participants. 
    The set of post-edited text is more stylistically homogeneous than the set of fully human-written text but less stylistically homogeneous than the set of \modelgenerated text. 
    }
    \Description{}
    \label{fig:between_particip}
\end{figure}

Here we analyze whether post-edited LLM text exhibits a consistent and identifiable style (H2). In \autoref{fig:between_particip}, we show the stylistic similarity between \modelgenerated text, text post-edited by different participants, and text written alone by different participants. Here, high stylistic similarity scores indicate a more homogeneous writing style within a group. For example, if all participants wrote their unassisted text in approximately the same style, the stylistic similarity between control texts from two random participants would be close to $1$.

We see that post-edited writing is indeed more stylistically homogeneous than unassisted writing (H2a, $p=.0002$, $g=1.42$, $95\%$ CI: [$1.33$, $1.51$]) and is less stylistically homogeneous than \modelgenerated writing (H2b, $p=.0002$, $g=-0.69$, $95\%$ CI: [$-0.74$, $-0.63$]). This suggests that post-editing injects a significant amount of stylistic diversity into \modelgenerated drafts, but not enough to reach the diversity of fully human-written text.

\begin{figure}[tb]
    \centering
    \includegraphics[width=0.95\linewidth, clip,trim=5 5 5 5]{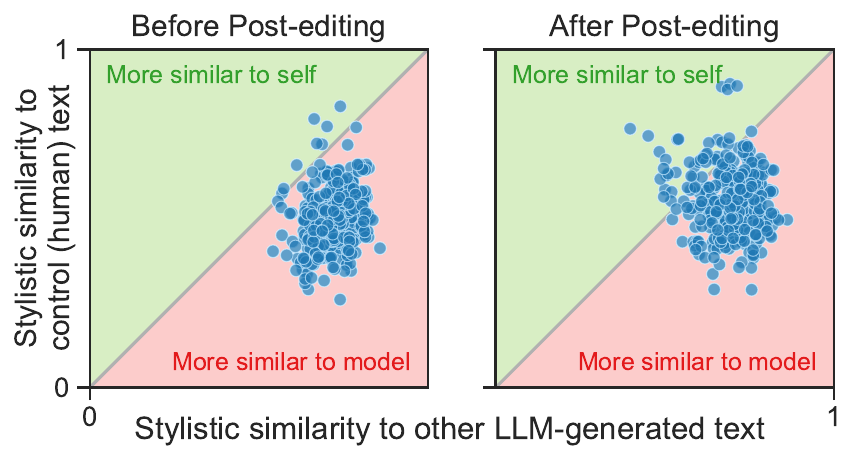}
    \caption{Similarity before (left) and after post-editing (right) to control and \modelgenerated text. Even after post-editing, text is more stylistically similar to \modelgenerated text than control text (H1c)}
    \label{fig:similarity_scatter_pre_post}
    \vspace{.5em}
    \centering
    \includegraphics[width=0.95\linewidth, clip,trim=5 5 5 5]{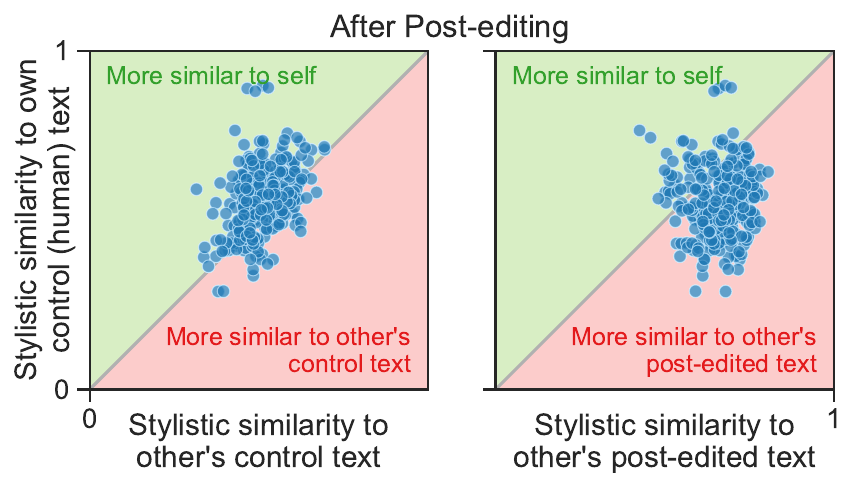}
    \caption{Similarity of post-edited text to participants' own control vs other participants' control (left) and post-edited (right) text. One's post-edited text is less similar to others' control text (related to H1a$^\prime$) and more similar to others' post-edited text (H2c).}
    \Description{}
    \label{fig:similarity_scatter_other_pe}
\end{figure}

However, in a similar pattern to H1c, participants' post-edited text is significantly more stylistically similar to other participants' post-edited text than to their own control text (H2c, $p=.0002$, $g=1.14$, $95\%$ CI: [$1.02$, $1.26$], \autoref{fig:similarity_scatter_other_pe} right) showing that some features of the stylistic fingerprint of \modelgenerated text are consistently preserved after post-editing across different participants. This result contrasts with our findings in H1a$^\prime$ in which we saw that post-edited writing is more similar to the post-editor’s own unassisted style than other participants’ \textit{unassisted} styles. Between H1a$^\prime$ and H2c, we can see that human-like changes made during post-editing are meaningfully unique to individuals while the AI-like aspects that are not addressed during post-editing are meaningfully shared between participants.

\subsection{Do Human Perceptions of Stylistic Similarity Agree with Model Measures?}\label{sec:h3}

We've observed that post-editing increases \textsc{LUAR}-measured stylistic-similarity to participants' unassisted writing and increases stylistic diversity but that text remains significantly more stylistically similar to \modelgenerated text than unassisted writing and significantly less diverse. However, \textsc{LUAR} embeddings may not perfectly translate to how participants experience their personal style.

\begin{figure}[tb]
\centering
    \includegraphics[width=.65\linewidth, clip,trim=6 7 7 6]{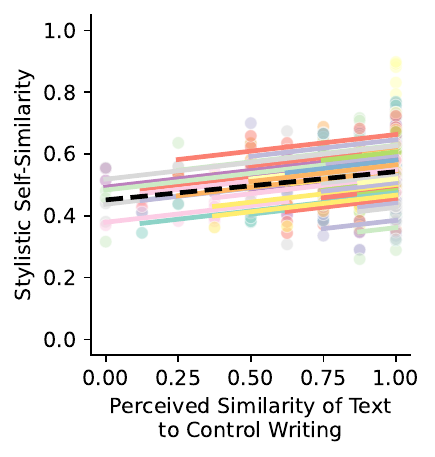}
    \caption{Relationship between participants' perception of and \texttt{LUAR} measurements of stylistic self-similarity for treatment text (H3). Y-axis is scaled from the original 1-5 range. Solid lines show the participant-level relationships from the repeated-measures correlation analysis and the black dashed line shows the overall correlation.}
    \Description{}
    \label{fig:perceived_vs_actual_self_similarity}    
\end{figure}

After writing or post-editing, participants were asked questions pertaining to how well each text captures their style which we average into a single score. Comparing this perceived stylistic similarity to \textsc{LUAR} stylistic similarity measurements, we find that they have a significantly positive but weakly calibrated correlation (H3, $r=0.244\pm0.076$, $p<.0001$, \autoref{fig:perceived_vs_actual_self_similarity}). %

\begin{figure}[tb]
\centering
    \includegraphics[width=.8\linewidth, clip,trim=2 2 2 2]{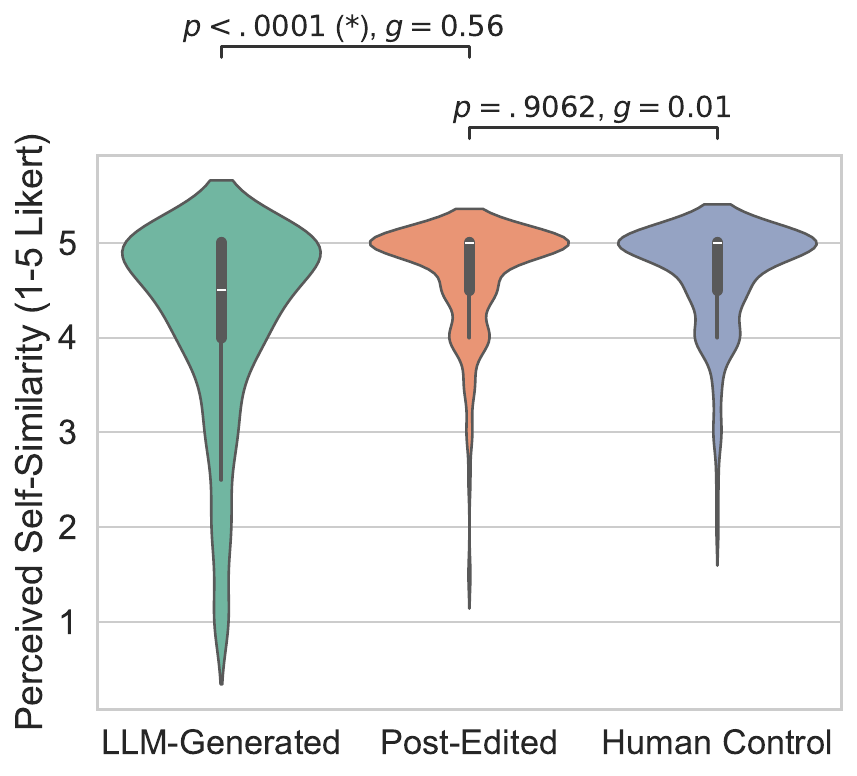}
    \caption{Differences in perceived self-similarity between \modelgenerated text, post-edited text, and control text.}
    \Description{}
    \label{fig:perceived_self_similarity}  
\end{figure}

We observed in H1c that, using \textsc{LUAR}-measured similarity, post-edited text was ultimately more stylistically similar to \modelgenerated text than participants' control text. However, while we've seen a significant correlation between the model measurements and human perceptions, we exploratorily observe that participants perceive their post-edited treatment text as no less representative of their personal style than their control text ($p=.9062$, $g=0.01$, $95\%$ CI: [$-0.17$, $0.19$], \autoref{fig:perceived_self_similarity}). This means that participants may not have experienced in practice the stylistic gap between their post-edited and control text measured by \textsc{LUAR} in H1c. They may not have been aware of certain LLM-associated features that they chose not to post-edit or the style embeddings may have included features that are not salient to participants' stylistic preferences in our studied writing tasks.

\subsection{What Are(n't) Participants Editing?}\label{sec:qual}

In an initial manual review, we observed substantial variation in the amount and density of edits to \modelgenerated text across participants. Some participants primarily edited the \modelgenerated text directly, while others instead added new text in their own style. As discussed in \autoref{sec:edit_len}, neither dense nor diffuse edits were more effective in increasing style-embedding-based similarity.

For further qualitative analysis, we open-coded $\sim20\%$ of the treatment documents. This included examples where the word error rate between LLM and final drafts was within one standard deviation of the mean, excluded cases with near-complete rewrites or minimal changes to better capture edit-level behavior, and prioritized more ``successful'' edits with larger increases in self-similarity.

These successful edits took a variety of forms. We observe many nonstandard edits, including potential typos and systematic differences in punctuation conventions, which may be accidental or reflect personal stylistic choices. We exploratorily analyze these less linguistically acceptable edits in \autoref{sec:appendix_grammar} and find that they increase stylistic similarity to control text, particularly the participant’s own writing, and more closely reflect individual-specific patterns rather than generic humanization.

Participants made edits reflecting regional and cultural norms, such as adopting British English conventions or adding explicitly religious language to eulogies and condolence letters. One participant reported inserting language from a ``traditional eulogy in the Church of England.'' These patterns suggest that certain groups may require more post-editing to express their personal style (e.g., to adjust spelling or add culturally specific language). For additional analysis of differences in post-editing effort across demographic groups, see \autoref{sec:H5}.

In addition, participants made fine-grained punctuation and lexical edits. As we discuss in \autoref{sec:lexical}, some edits removed words or punctuation commonly associated with \modelgenerated text (e.g., em dashes or ``dive'') or added contractions, while others reflected more idiosyncratic preferences, such as changing ``cozy'' to ``comfortable'' or ``assistance'' to ``help.'' While some edits could be interpreted as ``humanizing'', many instead reflected participants' personal sense of what sounded most like them. Some edits also had a semantic component: participants reported correcting factual content in $\sim31\%$ of treatment documents, and we observe related changes such as replacing ``Zoom'' with ``WhatsApp video call.''

Finally, participants sometimes simplified sentence structure or removed flowery language (e.g., changing ``an amazing spirit, one whose generosity knew no bounds'' to ``an amazing man''), while others added clich\'ed expressions such as ``She will forever be a part of our lives,'' suggesting this language was not uniformly viewed as uncharacteristic.

Overall, participants employed a diverse set of post-editing strategies to align \modelgenerated text with their personal writing style, highlighting the variety of features individuals find personally meaningful in their writing.

\subsection{(Why) Do Participants Like to Post-Edit?}\label{sec:why_postedit}

Here, we present exploratory results on the factors influencing participants' satisfaction with text written in this study and the process of post-editing. In our post-survey, participants were asked how they would prefer to write text in the future in scenarios where personal style is important. A majority ($58$) would prefer to post-edit text as they did in the treatment block, while a large contingency ($20$) would prefer to write without AI assistance. A few participants ($3$) would prefer to use model generations without post-editing. Overall, after participating in the study, participants reported being more likely to use AI in writing tasks where they believe capturing their personal style is important compared to their pre-study preferences ($t(80)=5.54$, $p<.0001$, $g=0.64$). Participants generally found writing alone somewhat more mentally demanding ($t(80)=-2.301$, $p=.024$, $g=-0.282$), time-consuming ($t(80)=-2.214$, $p=.03$, $g=-0.260$), and effortful ($t(80)=-2.602$, $p=.011$, $g=-0.266$) than post-editing. 

As we discuss in \autoref{sec:h1} and \autoref{sec:H4}, participants agreed with embedding-based measurements of stylistic similarity (which showed an increase in self-similarity after post-editing in \autoref{sec:h1}) and that increased perceived self-similarity was associated with higher perceived usability. Putting these findings together, we see that participants overall indicated that they believe their post-edited text matches their personal style better ($p<.0001$, $g=0.56$, $95\%$ CI: [$0.42$, $0.69$]) and is more usable ($p<.0001$, $g=0.53$, $95\%$ CI: [$0.39$, $0.66$]) than the original \modelgenerated text. 

In \autoref{fig:reason_rankings}, we compare the rankings of reasons why participants may prefer to post-edit \modelgenerated drafts or write alone. Participants preferring to write alone ranked the ``Originality'' of the \modelgenerated drafts as a somewhat more important factor than those preferring to post-edit. They ranked ``Efficiency'' and ``Reliability'' as less important. There were no significant differences between groups regarding the importance of ``Style'' (and the effort required to correct it). 
In other words, participants who prefer to write alone are less concerned with the effort of post-editing or the inclusion of details, and more with the perceived lack of originality in AI-generated drafts.

These exploratory results (expanded in \autoref{sec:appendix_future_pref}) suggest that potential adoption of post-editing may be limited less by the effort required (e.g., adjusting style) and more by a perceived lack of interesting or original content in \modelgenerated drafts.

\section{Conclusion}\label{sec:conclusion}

In this work, we examined people's ability to post-edit \modelgenerated text to capture their personal style in tasks where expressing said personal style is important. 
Post-editing significantly increased stylistic similarity to participants' own control text and decreased stylistic similarity to \modelgenerated text. Our findings show that participants can exert meaningful stylistic control through post-editing and are generally positive about using AI this way in tasks where expressing personal style matters.

However, despite these improvements, post-edited text remained measurably closer in style to \modelgenerated text than to participants' own fully human-authored text, and it exhibited reduced stylistic diversity relative to human control writing. Participants did not seem to perceive these residual stylistic markers: although their perception of stylistic self-similarity correlated with embedding-based measurements, they judged their post-edited text to be just as representative of their personal style as their control writing.

This gap between model-measured and participant-perceived stylistic similarity may have arisen for several reasons. First, participants may not have noticed certain LLM stylistic cues in their post-edited text, or they may have deemed these features irrelevant to their personal style. If participants failed to notice relevant cues, this may be caused by participants relying on simplified heuristics or algorithmic folk theories when judging whether text is AI-generated (e.g., associating ``delve'' or em dashes with LLM output), rather than attending to broader stylistic distributions. As a result, users (especially those with less hands-on experience writing with LLMs) may detect certain well-known cues while overlooking others. This interpretation is consistent with prior work showing that people develop partial mental models of algorithmic systems that are often incomplete or inaccurate~\citep[e.g.,][]{10.1145/3476046}, as well findings from \citet{russell-etal-2025-people} that suggest that frequent LLM users may develop stronger intuitions about AI writing style. 
Participants may easily gloss over and passively accept potential stylistic differences, rationalizing that a \modelgenerated draft is close enough to what they would have written alone, especially when they have a less developed sense of the space of stylistic differences to attend to.

\newpage %

Conversely, this gap may not reflect a lack of ``accuracy'' in participants' judgments, but rather a mismatch in what is being measured. Embedding-based metrics capture linguistic features that are statistically salient, but these features may not align with the aspects of style that lay users consider meaningful for their own writing. As a result, even highly ``accurate'' style embeddings may fail to predict how users perceive stylistic similarity. Differences in style judgments may not solely be a matter of user ability but also depend on what aspects of style are being focused on.

Given this potential gap between perceived and ``true'' stylistic similarity, future work could explore interventions that surface residual LLM stylistic features, allowing for more precise post-editing. However, these interventions may be of limited utility for users who already perceive their post-edited text as sufficiently representative of their personal style, and they could even encroach on users' autonomy by prescribing what it means to ``sound like'' themselves. Increasing the creativity and originality of initial LLM drafts may be a more impactful way to improve the utility of post-editing workflows, as generic or clich\'ed drafts could limit their effectiveness, even if they are easy to restyle.

Beyond post-editing workflows, potential ``inaccuracy'' in users' perceptions of their own style may be a concern for work focused on using LLMs to generate personalized text mimicking a user's style. In this setting, our findings suggest that users' intuitions about their own style may not be fully reliable and that additional scaffolding or feedback may be needed to help users make accurate judgments about style alignment.

Outside of our participants, it is unclear how audiences would perceive this post-edited text and whether would they notice cues participants overlooked, particularly in personal writing where audiences would be familiar with participants' style. Even when participants felt their post-edited text captured their personal style, LLM markers could undesirably influence how it is received.

In sum, our findings point to a disconnect between embedding-based representations of style and writers' experience of their personal style. Post-editing can support a sense of stylistic authenticity even when traces of LLM style remain. These results motivate further study of human-AI co-writing across settings, including how both writers and audiences interpret stylistic authenticity in AI-assisted personal writing.

\section{Limitations}

Throughout our study stylistic similarity (and AI detection) was measured via LLMs. While we select a style embedding model that is most performant in our domain and show that its output correlates with participants' self-assessments of their style, these measures may be less reliable than stylistic analyses conducted by human experts in forensic linguistics.

Stylistic self-similarity was measured using two unassisted reference texts which may not have captured all of a participant's personal stylistic preferences. Further, participants did not complete the same writing tasks in the control and treatment blocks, so the references used to measure stylistic self-similarity were collected on different writing tasks than the treatment texts. While some aspects of the participants' personal style are likely consistent between writing tasks, participants may have had task-specific stylistic preferences that were not captured in their control text. Overall, our measurements may have underestimated how well their treatment text captured participants' personal style and linguistic preferences.

While our study focused on real writing tasks where participants feel personal style is important, their writing is ultimately ``fake''. They are asked to not include real identifying details and the writing produced in the study will not ultimately be used for its intended purpose. Even if participants had a particular recipient in mind as they wrote an apology letter, this recipient will not be shown this apology. Participants' behaviors may be different when writing an apology letter with real details that will be sent to the real recipient.

We also acknowledge that our process for generating LLM drafts may not be representative of how real users would request a draft from an LLM. Users who do not, for example, provide bullet point details to an LLM may receive qualitatively different initial drafts and therefore have different experiences in post-editing. 

More broadly, the paradigm of an LLM providing an initial draft that users post-edit is of course not the only way for people and LLMs to write together~\cite{mysore-etal-2025-prototypical}. Some users may prefer to write the initial draft themselves and use an LLM for editing, some users may prefer more interaction turns, etc., and future work may consider how differently structuring the co-writing experience affects the style of the final text.

Our work considers participants' self-perceptions of the style and usability of the text they wrote or post-edited. However, this does not account for the perspective of the eventual reader. As we discussed in \autoref{sec:conclusion}, future work may examine whether audiences notice residual markers of AI style and how these influence their perception of the text, particularly for the writing tasks in which personal style is especially salient.

\section{Ethical considerations}

In our study, we encourage participants to provide fake details for the LLM to include in draft text. In a real-world setting, users may provide true details to an API-based LLM, introducing privacy risks. 

Our study does not consider whether or how participants feel their use of LLMs should be disclosed to their imagined recipient (in this case, that they planned and post-edited the text but did not write the full initial draft themselves). While user perceptions of their ownership of and contributions to the final text may vary~\citep[e.g.,][]{reza2025cowritingaihumanterms,10.1145/3711020}, obscuring the model's contributions via lack of disclosure or via removal of stylistic fingerprints lessens the reader's ability to identify ``who'' they are engaging with~\cite{10992536}. 

In this project, we used GitHub Copilot during implementation and ChatGPT while revising our (human-written) paper drafts.

\section*{Acknowledgments}
We sincerely thank the current and former members of the UMD CLIP lab for their valuable advice and feedback, especially Nishant Balepur, Yu Hou, and Navita Goyal as well as Jack Grieve.
We thank Pangram for providing API credits.
This research is supported in part by the Office of the Director of National Intelligence (ODNI), Intelligence Advanced Research Projects Activity (IARPA), via the HIATUS Program contract \#2022-22072200006. The views and conclusions contained herein are those of the authors and should not be interpreted as necessarily representing the official policies, either expressed or implied, of ODNI, IARPA, or the U.S. Government. The U.S. Government is authorized to reproduce and distribute reprints for governmental purposes notwithstanding any copyright annotation therein.
This material is based upon work partially supported by the NSF under Grant No. 2229885 (NSF Institute for Trustworthy AI in Law and Society, TRAILS).

\bibliography{custom, anthology}

\appendix

\section{Additional Experimental Details}\label{sec:overflow_exp}

\subsection{Participant Recruitment, Demographics, and Compensation}\label{sec:participant_demo}

We recruited $100$ participants\footnote{This is lower than planned in preregistration due to participants taking longer to complete our study than in our piloting. This change was made to achieve the expected hourly payment given our budget and was not based on study results.} for our study through the crowdsourcing platform Prolific.\footnote{\url{https://www.prolific.com/}} Participation was limited to one session per person and to English-fluent users, as non-fluent speakers may have a less clear sense of personal style in English. Participants took an average of $1.69$ hours to complete the study and were compensated at US\$15 per hour on average including a US\$$3$ ``bonus''. To discourage low-effort submissions participants were told they would receive the bonus only if their style was consistent throughout the study. In practice, all participants were paid the full amount. 
Following our preregistered exclusion criteria, we discarded responses where participants \textit{never} edited any of the \modelgenerated drafts or did not complete the entire study, leaving $81$ participants.

In our post-survey, all demographic questions were optional, and participants could select more than one gender and race category. $67.9\%$ of these participants self-identified as women and $32.1\%$ as men, one of which identified as a transgender man. $23.5\%$ of participants were between the ages of $18$-$25$, $45.7\%$ between $26$-$40$, $18.5\%$ between $41$-$60$, and $3.7\%$ over the age of $60$ with $8.6\%$ preferring not to report.
$74.1\%$ of participants self-identified as Black or African American, $13.6\%$ as White, $4.9\%$ as Asian, and $1.2\%$ as Native Hawaiian or Other Pacific Islander with $6.2\%$ preferring not to report. $3.7\%$ of participants self-identified as Hispanic or Latino. $1.2\%$ of participants reported their highest level of educational achievement to be a High school diploma or equivalent, $1.2\%$ as Some college, no degree, $2.5\%$ as Associate degree, $6.2\%$ as Bachelor's degree, $1.2\%$ as Master's degree, $55.6\%$ as Professional degree, $18.5\%$ as Doctorate, and  $13.6\%$ as Other.

\subsection{Permutation Test Details}\label{sec:permutation_tests}

Much of our main analysis was conducted using two-tailed permutation tests with $10,000$ permutations. For example, in H1a, we consider whether participants' treatment text sounds more like their control writing after post-editing than before. Using a style embedding model, we generate an embedding of each participant's concatenated control texts, an embedding for each treatment text before post-editing, and an embedding for each treatment text after post-editing. Taking the cosine similarity between treatment and control embeddings, we then have a measurement for the treatment texts' stylistic similarity to the participant's control text before and after they post-edited the treatment text. We calculate the mean similarity of each list and treat the absolute difference as the observed mean. We then consider $10,000$ permutations of these scores. We can then calculate $p$-value as the percentage of permutations where the mean absolute difference between groups is greater than the observed mean.

Where applicable, we match the writing tasks in reference texts between comparisons. For example, for a single participant, we collect two unassisted control texts. When we calculate similarity of that participant's treatment text to \modelgenerated text, we only consider comparisons to \modelgenerated text that was generated for the same writing task as that participant's control tasks.

\begin{table*}
    \footnotesize
    \centering
    \begin{tabular}{m{5.2cm}m{5.2cm}m{2.5cm}m{1.1cm}}
        \toprule
        Original \modelgenerated Text & Post-edited Text& \multicolumn{2}{l}{Comparison Measures}  \\
        \midrule
        Dear Jennifer,\newline\newline I’m so sorry for my \DIFdel{behavior} at the Italian restaurant last night. I shouted at you because I felt you dismissed my worries about my job and I believed you were being insincere. I was under a lot of stress and pressure, and I let my emotions get the better of me. I \DIFdel{realize} now that I should have communicated my feelings calmly instead of reacting \DIFdel{impulsively}.\newline\newline I know you were only trying to be supportive and caring, and I truly appreciate how hard \DIFdel{you’ve} listened and stood by me. My reaction was unfair to you and hurtful. I regret causing \DIFdel{you pain.}\newline\newline Our friendship means the world to me, and I hope you can forgive me. I am committed to making things right and doing what\DIFdel{ever} it takes to rebuild your trust. Thank you for being there, and I \DIFdel{hope we can move forward together}.\newline\newline Sincerely,\newline Your friend & 
        Dear Jennifer,\newline\newline I’m so sorry for my \DIFadd{behaviour} at the Italian restaurant last night. I shouted at you because I \DIFadd{(wrongly) }felt you dismissed my worries about my job and I believed you were being insincere. I was under a lot of stress and pressure, and I let my emotions get the better of me. I \DIFadd{realise} now that I should have communicated my feelings calmly\DIFadd{,} instead of reacting \DIFadd{as I did}.\newline\newline I know you were only trying to be supportive and caring, and I truly appreciate how hard \DIFadd{you have} listened and stood by me. My reaction was unfair to you and hurtful. I regret \DIFadd{upsetting you and} causing \DIFadd{a bit of a scene in the restaurant.- to put it mildly!}\newline\newline Our friendship means the world to me, and I hope you can forgive me. I am committed to making things right and doing what it takes to rebuild your trust \DIFadd{and friendship.}. Thank you for being there\DIFadd{ in the past}, and I\DIFadd{ really hope you will be there in the future- as I will also be there for you.}.\newline\newline Sincerely,\newline Your friend\newline& \{\newline\phantom{--}$\Delta$ self-sim: \newline \phantom{--}$\Delta$ LLM-sim (avg): \newline \phantom{--}$\Delta$ Pangram Score: \newline \phantom{--}WER: \newline \} & \phantom{$-$}$0.202$,\newline $-0.197$,\newline $-0.999$,\newline \phantom{$-$}$0.320$    \\
        \rowcolor{customrowcolor}My \DIFdel{love}, today I stand before you grateful for every laugh, every adventure, every quiet moment shared. Your kindness, patience, and \DIFdel{quick wit} have made me stronger, braver, and joyful. I promise to support you in your dreams, to celebrate your successes, and to lift you up when challenges arise. I vow to be your partner in all \DIFdel{of Life’s journeys}, from road trips at dawn to cozy nights on the couch. I promise to share the remote. I pledge to listen with understanding, to forgive with compassion, and to always choose love over pride. Together we will build a home filled with laughter, honesty, and \DIFdel{curiosity}. I cherish our memories—the picnic that turned into a dance, our first sunrise hike—and look forward to creating many more. With sincerity and devotion, I commit my heart to you, now and always, as we grow, learn, and chase our dreams side by side.  & My \DIFadd{soon to be husband}, today I stand before you grateful for every laugh, every adventure, every quiet \DIFadd{and loud }moment\DIFadd{s} shared. Your kindness, patience, and \DIFadd{love} have made me stronger, braver, and joyful. I promise to support you in your dreams, to celebrate your successes, and to lift you up when challenges arise. I vow to be your partner in all \DIFadd{the journey of life}, from road trips at dawn to cozy nights on the couch. I promise to share the remote. I pledge to listen with understanding, to forgive with compassion, and to always choose love over pride. Together we will build a home filled with laughter, honesty, and \DIFadd{love} I cherish our memories—the picnic that turned into a dance, our first sunrise hike—and look forward to creating many more. With sincerity and devotion, I commit my heart to you, now and always, as we grow, learn, and chase our dreams side by side. & \{\newline\phantom{--}$\Delta$ self-sim: \newline \phantom{--}$\Delta$ LLM-sim (avg): \newline \phantom{--}$\Delta$ Pangram Score: \newline \phantom{--}WER: \newline \} & $-0.029$,\newline $-0.007$,\newline \phantom{$-$}$0.000$,\newline \phantom{$-$}$0.093$    \\
        \bottomrule
    \end{tabular}
    \caption{Example \modelgenerated and post-edited text from our study. The first is a relatively ``successful'' case where the \textsc{LUAR}-measured similarity to the participant's unassisted control text increases and the \textsc{LUAR}-measured similarity to other \modelgenerated text as well as the Pangram-predicted AI-likelihood decrease through post-editing. The second is less successful with the \textsc{LUAR}-measured similarity to the participant's unassisted control text not improving despite the edits made. For the full set of text written, generated, or post-edited in the treatment and control blocks of our study, see \url{https://github.com/ctbaumler/personal_style_postedit}.}
    \label{tab:example_output}
\end{table*}

\section{Additional Analyses}\label{sec:other_results}

Here we report additional analyses and expanded details from analyses in \autoref{sec:quant_results}. These results include discussion of three additional preregistered hypotheses.

\begin{description}[nolistsep,noitemsep]
    \item[H4:] In writing tasks where participants consider authentic style to be important, they are more likely to report willingness to use a piece of text they deem to match their style more accurately.
\end{description}

Beyond individualized style, we consider how post-editing labor is distributed among people from various demographic backgrounds who value writing text that communicates this community membership. 

\begin{description}[nolistsep,noitemsep]
    \item[H5:] Participants from marginalized communities who want that community membership to come across in their writing will require more post-editing to capture their personal style.
\end{description}

H4 and H5 are discussed in \autoref{sec:H4} and \autoref{sec:H5} respectively. We confirm that the style of \modelgenerated text used in our study is consistent across participants (\autoref{sec:llm_consistency}) and discuss our choice of style embedding model (\autoref{sec:emb_selection}). We provide additional details about our analysis of the reasons why our participants would or would not prefer to post-edit from \modelgenerated drafts in future writing (\autoref{sec:appendix_future_pref}). We also explore whether our results in H1 change on tasks where personal style is viewed as somewhat less important (\autoref{sec:style_importance}) and various aspects of participants' edits including their lexical choices, length, etc (\autoref{sec:appendix_edits}). 

\begin{table*}[tb]
    \centering
    \begin{tabular}{l c c c }
    \toprule
    Model  & MRR ($\uparrow$) & R@1 ($\uparrow$) & R@8 ($\uparrow$)\\
    \midrule
    \citet{rivera-soto-etal-2021-learning} (\href{https://huggingface.co/rrivera1849/LUAR-MUD}{\texttt{LUAR-MUD}}) & 0.5888 & 0.4506 & 0.8333\\
    \citet{rivera-soto-etal-2021-learning} (\href{https://huggingface.co/rrivera1849/LUAR-CRUD}{\texttt{LUAR-CRUD}}) & 0.4966 & 0.3519 & 0.7778\\
    \citet{kim-etal-2025-leveraging} (\href{https://huggingface.co/Blablablab/multilingual-style-representation}{\texttt{multilingual-style-representation}}) & 0.4364 & 0.2901 & 0.6975\\
    \citet{wegmann-etal-2022-author} (\href{https://huggingface.co/AnnaWegmann/Style-Embedding}{\texttt{CISR}}) & 0.4058 & 0.2778 & 0.6481\\
    \citet{patel-etal-2025-styledistance} (\href{https://huggingface.co/StyleDistance/styledistance}{\texttt{StyleDistance}}) & 0.3688 & 0.2469 & 0.6049\\
    \citet{Koornstra2023} (\href{https://huggingface.co/TimKoornstra/SAURON}{\texttt{SAURON}}) & 0.3457 &  0.2284 & 0.5000\\
    \midrule
    Random & 0.0615 & 0.0123 & 0.0988\\
    \bottomrule
\end{tabular}
\caption{Authorship identification performance of style embedding models on our control text. Authorship rankings are made with $80$ references from other participants and one reference from the true author. Both versions of \texttt{LUAR} are distributed under the terms of the Apache License (Version 2.0). \texttt{StyleDistance} is distributed under the terms of an MIT license. \texttt{SAURON} is distributed under the terms of the GNU General Public License v3.0.}
    \label{tab:ta2}
\end{table*}

\subsection{\modelgenerated Text's Style is Consistent Between Participants}\label{sec:llm_consistency}

As we discussed in \autoref{sec:study_design}, we provide the LLM generating treatment drafts with an initial set of planned details provided by the participant. For example, if the task is to write a an apology letter, the participant might provide bullet points summarizing what they did wrong and the steps they will take to avoid repeating the problem. 
These bullet points may include some markers of participants' personal style. While the model was not prompted to do so, it is possible that the model could pick up on some of these markers and use them when creating a draft. This would mean that participants were not post-editing from a ``generic'' or ``default'' LLM style but a somewhat personalized style. To confirm that this did not happen, we consider the stylistic consistency between model generations produced for the same participant or for different participants (in a procedure similar to that done for H2a and H2b in \autoref{sec:h2}). If the model is taking the style of the bullet points into account, we would expect to see higher stylistic similarity between text generated using the same participant's bullet points than we do between text generated using different participant's bullet points. We observe no such difference ($p=.697$). 

\subsection{Style Embedding Model Selection}\label{sec:emb_selection}

\begin{table*}[tb]
    \centering
    \begin{tabular}{lclccl}
    \toprule
    Hypothesis & $p$ & & $g$ & $95\%$ CI & Model\\
    \midrule
    \multirow{2}{4em}{H1a} & $0.0002$ & $*$ & $\phantom{-}0.55$ & [$\phantom{-}0.38$, $\phantom{-}0.71$] & \texttt{LUAR-MUD}\\
     & $0.0002$ & $*$ & $\phantom{-}0.48$ & [$\phantom{-}0.32$, $\phantom{-}0.64$] & \texttt{CISR} \\
    \rowcolor{customrowcolor} & $0.0002$ & $*$ & $-0.56$ & [$-0.70$, $-0.43$] & \texttt{LUAR-MUD}\\
    \rowcolor{customrowcolor} \multirow{-2}{4em}{H1a$^\prime$} & $0.0002$ & $*$ & $-0.30$ & [$-0.44$, $-0.16$] & \texttt{CISR} \\
    \multirow{3}{4em}{H1b} & $0.0002$ & $*$ & $-0.41$ & [$-0.44$, $-0.39$] & \texttt{LUAR-MUD}\\
     & $0.0002$ & $*$ & $-0.55$ & [$-0.57$, $-0.53$] & \texttt{CISR} \\
     & $0.0002$ & $*$ & $-0.45$ & [$-0.55$, $-0.35$] & Pangram \\
    \rowcolor{customrowcolor} & $0.0002$ & $*$ & $-1.43$ & [$-1.55$, $-1.32$] & \texttt{LUAR-MUD}\\
    \rowcolor{customrowcolor}\multirow{-2}{4em}{H1c} & $0.0002$ & $*$ & $-2.30$ & [$-2.46$, $-2.15$] & \texttt{CISR} \\
    \multirow{2}{4em}{H2a} & $0.0002$ & $*$ & $\phantom{-}1.42$ & [$\phantom{-}1.33$, $\phantom{-}1.51$] & \texttt{LUAR-MUD}\\
     & $0.0002$ & $*$ & $\phantom{-}1.03$ & [$\phantom{-}0.91$, $\phantom{-}1.14$] & \texttt{CISR} \\
    \rowcolor{customrowcolor} & $0.0002$ & $*$ & $-0.69$ & [$-0.74$, $-0.63$] & \texttt{LUAR-MUD}\\
    \rowcolor{customrowcolor} \multirow{-2}{4em}{H2b}& $0.0002$ & $*$ & $-0.66$ & [$-0.70$, $-0.62$] & \texttt{CISR} \\
    \multirow{2}{4em}{H2c} & $0.0002$ & $*$ & $\phantom{-}1.14$ & [$\phantom{-}1.02$, $\phantom{-}1.26$] & \texttt{LUAR-MUD}\\
     & $0.0002$ & $*$ & $\phantom{-}2.09$ & [$\phantom{-}1.94$, $\phantom{-}2.24$] & \texttt{CISR} \\
    \bottomrule
    \end{tabular}
    \caption{Comparison of H1 and H2 results between style embedding (and AI detector) models. We see that the significance of effects and the direction of effect sizes are the same regardless of model choice though the magnitude of the effect sizes somewhat vary. }
    \label{tab:p_h1/2}
\end{table*}

We select a style embedding model to use in our analysis from a set of six models~\cite{kim-etal-2025-leveraging, rivera-soto-etal-2021-learning, wegmann-etal-2022-author, Koornstra2023, patel-etal-2025-styledistance}. As these models' robustness may vary based on factors such as the domain of the text samples, their length, etc, we choose to evaluate them on an authorship identification task on our control data. Using this setup allows us to choose the model that best captures salient stylistic features in our study's writing tasks.

In the authorship identification task, the model is given a query document written by an unknown author and set of reference documents. The task is then to rank the reference documents based on how likely they are to be written by the same author as the query document. One of these reference documents is written by the same author as the query document and ought to be ranked first. 

Say we are considering the first control document from a participant as a query. To construct the set of reference documents, we first take the participant's other control document as the correct reference. For every remaining participant, we include one of their two control documents either selecting at random or selecting the document that was not written for the same task as the query. For each model, we embed the query document and each of the $81$ reference documents. The final predicted ranking of reference authors is calculated using the cosine similarity between the query embedding and each reference document embedding. 

In \autoref{tab:ta2}, we show the mean reciprocal rank, average recall at 1 ($1$ if the true author is ranked first), and average recall at 8 ($1$ if the true author is ranked in the top $8$) for each model as well as for random rankings. We observe that all models perform much better than random ranking. \texttt{LUAR-MUD} performs best on this proxy task, so we use this model's embeddings for the rest of our analysis. We also repeat our analysis for H1 and H2 using \texttt{CISR}. As we see in \autoref{tab:p_h1/2}, the conclusions between these models are the same, though the effect sizes sometimes differ in magnitude.

\subsection{Humans Perceptions of Stylistic Authenticity vs Usability of Text}\label{sec:H4}

\begin{figure}[tb]
    \centering
    \includegraphics[width=0.7\linewidth, clip,trim=6 7 7 6]{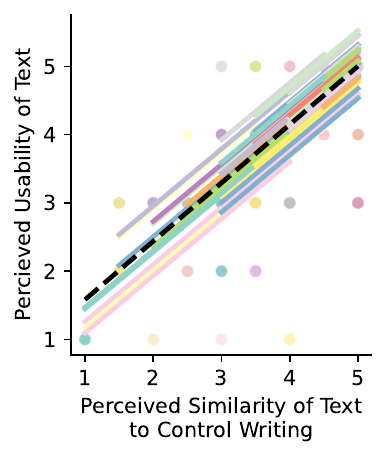}
    \caption{Relationship between participants' perception of the usability and stylistic self-similarity of treatment text. Solid lines show the participant-level relationships from the repeated-measures correlation analysis and the black dashed line shows the overall correlation.}
    \label{fig:perceived_usability}
\end{figure}

As we discuss in \autoref{sec:formative}, our study includes writing tasks where formative study participants (and our main study participants) believe expressing their personal style is important. This should mean that participants will think a text is more usable (e.g., they are more likely to send someone the apology they wrote) if it is stylistically authentic (H4). However, there is a potential for mismatch between how important participants \textit{claim} style is and how much style affects their \textit{perceived usability} of text. In other words, people may claim that style is important to writing an apology but, in practice, may still be willing to send an apology letter that does not match their true style, echoing concerns from other human-AI tasks about mismatch between user's stated preferences and their behavior.~\cite{balepur2025goodplanhardfind, balepur-etal-2024-smart, wen2024languagemodelslearnmislead, mozannar2024realhumanevalevaluatinglargelanguage}

However, we see that, for the writing tasks considered in this study, participants say their are more likely to use text when they believe it is more stylistically authentic (H4, $r=0.774\pm0.031$, $p<.0001$, \autoref{fig:perceived_usability}). We note that this usability measurement is based on a self-reported willingness to use the text written or post-edited in this study in it's intended purpose. It is possible that participants' observed behaviors would not match these measurements. That is, this result does not rule out the possibility of observing a disconnect between perceived stylistic authenticity and a user's decision to use or not use text in practice.

\begin{table*}[tb]
\centering
\small
\begin{tabular}{lrrrrr}
\toprule
Predictor & Estimate & SE & z & p & 95\% CI \\
\midrule
Intercept & 0.507 & 0.008 & 61.427 & <.001 & [0.490, 0.523] \\
post-edited (True) & 0.055 & 0.006 & 9.167 & <.001 & [0.043, 0.066] \\
personal style importance (Non-max) & -0.010 & 0.011 & -0.975 & .329 & [-0.031, 0.010] \\
post-edited (True) $\times$ personal style importance (Non-max) & -0.017 & 0.013 & -1.280 & .201 & [-0.042, 0.009] \\
\bottomrule
\end{tabular}
\caption{Parameter estimates from a mixed-effects model predicting stylistic self-similarity from post-editing and the perceived importance of personal style.}
\label{tab:h1a_importance}
\phantom{-}

\begin{tabular}{lrrrrr}
\toprule
Predictor & Estimate & SE & z & p & 95\% CI \\
\midrule
Intercept & 0.730 & 0.004 & 184.014 & <.001 & [0.722, 0.737] \\
post-edited (True) & -0.039 & 0.001 & -41.097 & <.001 & [-0.041, -0.037] \\
personal style importance (Non-max) & -0.020 & 0.002 & -11.474 & <.001 & [-0.024, -0.017] \\
post-edited (True) $\times$ personal style importance (Non-max) & 0.020 & 0.002 & 9.614 & <.001 & [0.016, 0.024] \\
\bottomrule
\end{tabular}
\caption{Parameter estimates from a mixed-effects model predicting stylistic similarity to \modelgenerated text from post-editing and the perceived importance of personal style.}
\label{tab:h1b_importance}
\phantom{-}

\begin{tabular}{lrrrrr}
\toprule
Predictor & Estimate & SE & z & p & 95\% CI \\
\midrule
Intercept & 0.694 & 0.006 & 125.698 & <.001 & [0.683, 0.705] \\
same-participant (True) & -0.125 & 0.005 & -25.670 & <.001 & [-0.135, -0.116] \\
personal style importance (Non-max) & -0.009 & 0.002 & -4.389 & <.001 & [-0.013, -0.005] \\
same-participant (True) $\times$ personal style importance (Non-max) & -0.053 & 0.011 & -4.977 & <.001 & [-0.073, -0.032] \\
\bottomrule
\end{tabular}
\caption{Parameter estimates from a mixed-effects model predicting stylistic similarity from reference source (\modelgenerated vs human-written) and the perceived importance of personal style.}
\label{tab:h1c_importance}
\end{table*}

\subsection{Inequities in Effort to Post-Edit to Capture Individual Style Across Demographic Groups}\label{sec:H5}

\begin{figure}[tb]
    \centering
    \includegraphics[width=0.99\linewidth]{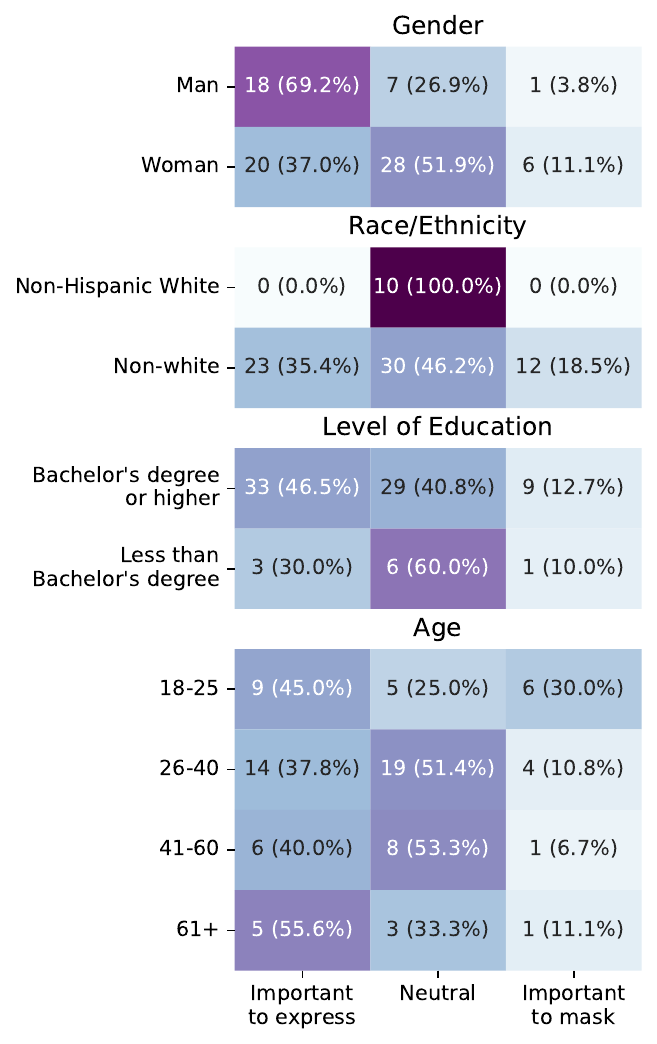}
    \caption{Comparison of participant's desires for various aspects of their identity to be expressed or masked in the writing tasks they completed in our study. Percentages are computed at the row-level.}
    \label{fig:demographic_expression}
\end{figure}

We hypothesized (H5) that there may be unequal burden in post-editing to capture individual style between participants from marginalized communities who want to express this facet of their identity in their writing and participants from privileged groups. However, we find that in our study, the slices of data available for this planned analysis were relatively small. While (as we discussed in \autoref{sec:emb_selection}), we saw consistency between style embedding models in H1 and H2 (which consider our full dataset), stylistic self-similarity as measured by \texttt{LUAR} and \texttt{CISR } are not always consistent in the relevant slices of data.

For example, our study included $10$ non-Hispanic white participants\footnote{Analyses in this section use bucketed demographic groups due to the small overall sample size (e.g., we grouped the single transgender male participant with the cisgender men). Participants with fully or partially unspecified answers were not included (e.g., we had one cisgender participant with unspecified gender).} and $23$ Hispanic or non-white participants who said it was important that their race/ethnicity come across in the writing tasks in our study. Using \texttt{CISR} to measure style, we find that non-Hispanic white participants had a higher stylistic similarity between initial \modelgenerated text and their unassisted control text ($p=.0002$, $d=1.08$, $95\%$ CI: [$0.73$, $1.52$]), meaning that participants from a privileged background would need less post-editing to express their personal style. However, using \texttt{LUAR} to measure style, we observe the opposite effect ($p=.0162$, $d=-0.54$, $95\%$ CI: [$-0.91$, $-0.18$]). For non-Hispanic white participants ($p=0.1001$, $r=-1.20$) and for Hispanic or non-white participants who want to express this identity ($p=0.0009$, $r=-0.26$) we see respectively insignificant and significantly negative repeated measures correlation between the initial self-similarity scores produced by both models. However, the overall repeated measures correlation between all self-similarities of all participants was significantly positive ($p<.0001$, $r=0.39$). This shows that the style embedding models used in this analysis may not be sensitive enough to demographic-associated stylistic differences to reliably these measure differences in small slices of participants.

Participants from different demographic groups varied in whether they wanted these aspects of their identity to come across in their personal writing (\autoref{fig:demographic_expression}). Exploratory analysis suggested that the desire to express gender in personal writing varied somewhat between men and women ($\chi^2(2) = 7.38$, $p = .025$, $\text{Cram\'er's V} = .30$), with men more often reporting a desire for their gender to come across. Similarly, the desire to express race and ethnicity varied between non-Hispanic white and non-white participants ($\chi^2(2) = 10.10$, $p = .006$, $\text{Cram\'er's V} = .37$), with non-Hispanic white participants being neutral and non-white participants exhibiting more varied preferences, including wanting their race or ethnicity either expressed or masked. No exploratory evidence indicated variability in the desire to express one's level of education or age.

Our post-survey did not consider factors that may influence participants' preferences, such as the intended audience or type of writing task. Preferences may differ for more formal writing contexts beyond the tasks in this study.

\subsection{Effectiveness of Post-Editing When Personal Style is Less Important}\label{sec:style_importance}

As we discussed in \autoref{sec:h1}, we see that post-editing significantly increased stylistic similarity to participants' control text (H1a) and significantly decreased stylistic similarity to \modelgenerated text (H1b) with the post-edited text ultimately being significantly more LLM-like than self-similar (H1c). In \autoref{sec:formative}, we show that the tasks used in this study are ones where participants largely find expressing personal style to be important. However, there was some variability with participants reporting the maximum likert score of stylistic importance in $79\%$ of the tasks they completed in treatment blocks (see \autoref{fig:task_importance} for the full distribution over both treatment and control tasks). 

Here we explore whether the effects observed in H1 change when participants complete writing tasks where they believe personal style is less important. We binarize participants' task-level likert ratings into ``max'' (a $5$ on the $5$-point scale) and ``non-max'' (below a $5$ with an average of $3.8$ for treatment tasks). Then we fit three mixed effect models corresponding to H1a, H1b, and H1c.

For H1a, the mixed effect model predicts the stylistic similarity between a treatment document and a control human-authored reference, varying whether the treatment text is post-edited and the importance of style for a given participant on the current task (see \autoref{tab:h1a_importance}). This exploratory analysis found no evidence that the importance of style affects the effectiveness of post-editing in changing self-similarity.

For H1b, the mixed effect model predicts the stylistic similarity between a treatment document and an \modelgenerated reference, varying whether the treatment text is post-edited and the importance of style for a given participant on the current task (see \autoref{tab:h1b_importance}). Here we exploratorily observe that style importance may significantly moderate the effect of post-editing ($\beta = 0.020$, $SE = 0.002$, $z = 9.614$, $p<.001$). When participants post-edited text for a task where they believe personal style is more important, post-editing reduced model-similarity more strongly. Participants may have felt less of a need to remove LLM stylistic markers when they felt less of a need to express their personal style.

For H1c, the mixed effect model predicts the stylistic similarity of a post-edited document, varying whether the similarity is measured to a human-written control reference or an \modelgenerated reference and the importance of style for a given participant on the current task (see \autoref{tab:h1c_importance}). Here we observe exploratorily that style importance may significantly moderate the gap in LLM- and self-similarity ($\beta = -0.053$, $SE = 0.011$, $z = -4.977$, $p<.001$). Text that was post-edited for a task where the participant believed personal style is more important was more similar to \modelgenerated text. This again suggests that participants may have felt less of a need to remove LLM stylistic markers when they felt less of a need to express their personal style.

Overall, in this exploratory analysis we observe some evidence that participants' post-edits were less effective in removing stylistic features characteristic of \modelgenerated text when they believed the writing task was one where personal style was less important. We note that the variability in style importance was fairly low in our study. These effects may be more prominent when comparing to tasks where personal style is even less important to users such as professional or formal writing.

\subsection{Reasons (Not) to Post-Edit}\label{sec:appendix_future_pref}

\begin{table}[tb]
    \centering
    \footnotesize
    \begin{tabular}{lcc}
        \toprule
        Reason & $p$ & Cliff's $\delta$ \\
        \midrule
        Originality & 0.011 & \phantom{-}0.375 \\ 
        Privacy & 0.103 & \phantom{-}0.243 \\ 
        Overall Quality & 0.479 & \phantom{-}0.106 \\ 
        Style & 0.532 & \phantom{-}0.094 \\ 
        Ownership & 0.668 & -0.065 \\ 
        Efficiency & 0.006 & -0.415 \\ 
        Reliability & 0.004 & -0.426 \\ 
        \bottomrule
    \end{tabular}
    \caption{Exploratory Mann-Whitney U tests comparing feature prioritization from participants who would prefer to post-edit vs write alone in the future. Features with higher relative ranking for post-editing include Efficiency and Reliability, while Originality shows higher relative ranking among participants who would prefer to write alone.}
    \label{tab:reason_ranking}
\end{table}

\begin{figure}[tbp]
    \centering
    \includegraphics[width=0.9\linewidth]{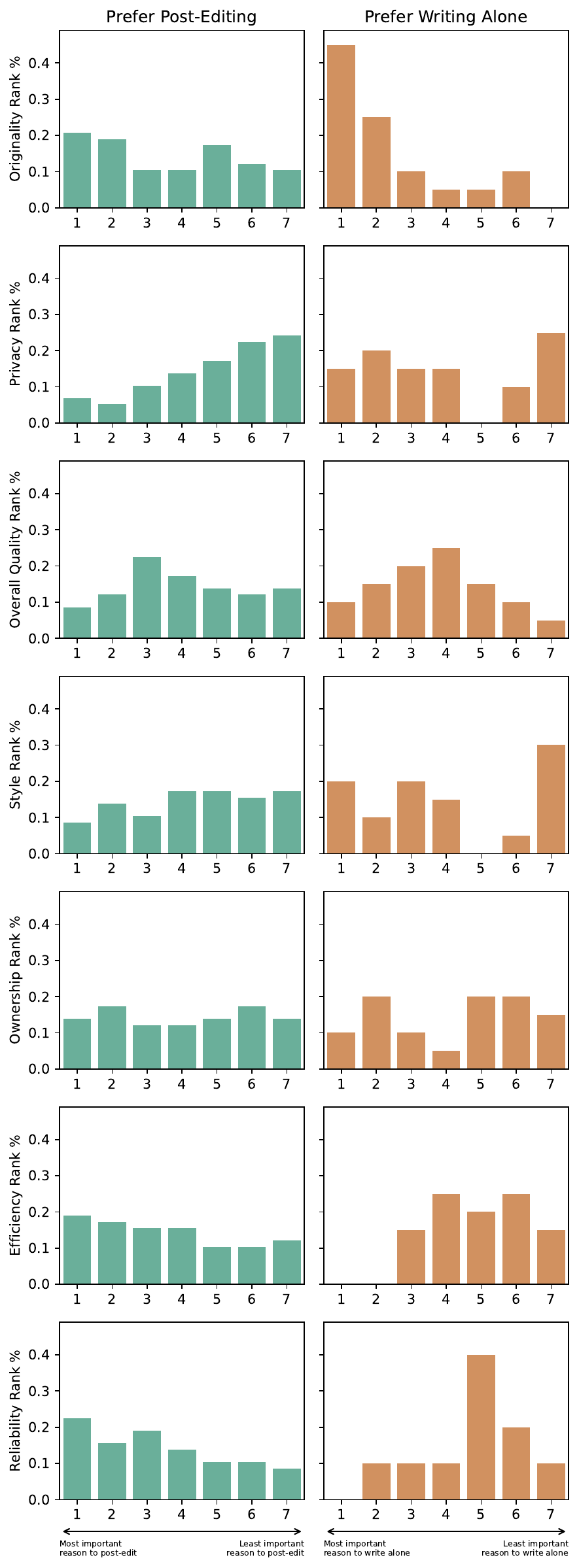}
    \caption{Comparison of feature prioritization between participants who would prefer to post-edit or write alone in the future.  See \autoref{tab:reason_ranking} for Mann-Whitney U test results.}
    \label{fig:reason_rankings}
\end{figure}

\begin{figure*}[ht!]
    \centering
    \includegraphics[width=.9\textwidth,clip,trim=240 1910 240 0]{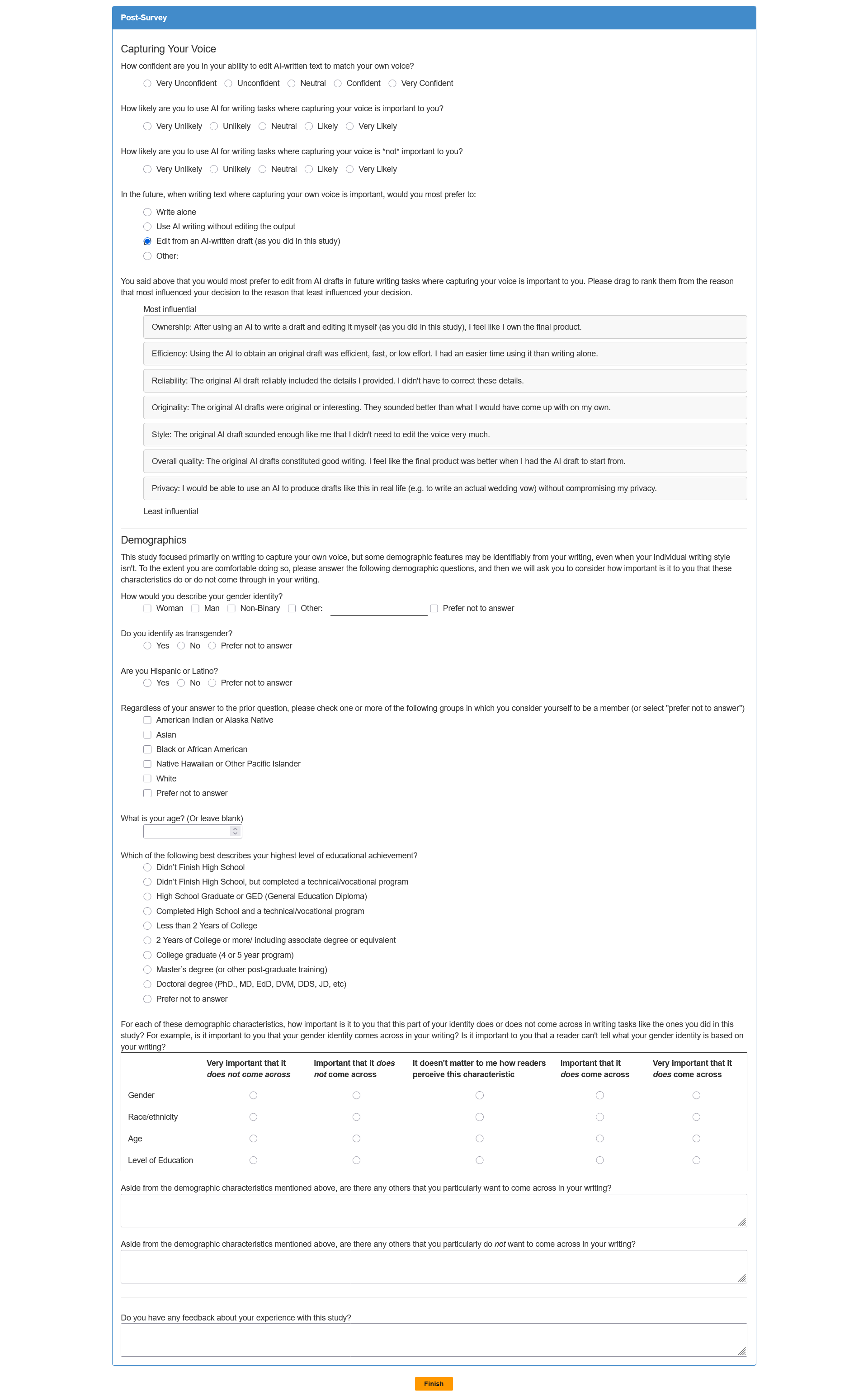}
    \caption{First half of the post-study survey. In the interface, both these questions and those in \autoref{fig:post_survey_2} appear on a single screen.}
    \label{fig:post_survey_1}
    \Description{}
\end{figure*}

As discussed in \autoref{sec:why_postedit}, many participants reported a desire to post-edit from \modelgenerated drafts in similar writing tasks in the future, while others preferred to write alone. In the post-survey, participants ranked the reasons influencing this decision from most to least important (see \autoref{fig:post_survey_1} for a full list with descriptions). Participants who preferred to post-edit were asked to rank positively framed reasons (e.g., ``The original AI drafts were original or interesting. They sounded better than what I would have come up with on my own.''), while participants who preferred to write alone ranked negatively framed reasons (e.g., ``The original AI drafts were too clich\'e or unoriginal. They weren't any more interesting or original than what I would have come up with alone.''). These framing differences allowed us to probe the benefits that lead participants to want to post-edit and the barriers that lead participants to want to write alone. 

\autoref{fig:reason_rankings} shows the distribution of feature rankings for participants who would prefer to post-edit versus write alone (excluding the three participants who would use \modelgenerated text as-is). In \autoref{tab:reason_ranking}, we report exploratory Mann–Whitney U tests comparing these distributions, along with Cliff's $\delta$ effect sizes.

We find some evidence that ``Efficiency'' and ``Reliability'' are ranked as relatively more important by participants who would prefer to post-edit. Participants who valued the convenience of using the LLM tended to want to use it again. Conversely, participants who prioritized ``Originality'' were somewhat more likely to prefer writing alone. If \modelgenerated drafts were perceived as unoriginal or clich\'ed, this may reduce their utility as an initial draft. We observe no evidence of group differences in the relative importance of ``Style'', ``Ownership'', or ``Overall Quality''. This suggests that effort to make \modelgenerated text feel authentic and any reduced sense of ownership did not appear to be major barriers to future use.

\subsection{Characterizing Post-Editing Behavior}\label{sec:appendix_edits}

\subsubsection{Linguistic Acceptability of Edits}\label{sec:appendix_grammar}

\begin{figure}[tb]
    \centering
    \includegraphics[width=0.9\linewidth]{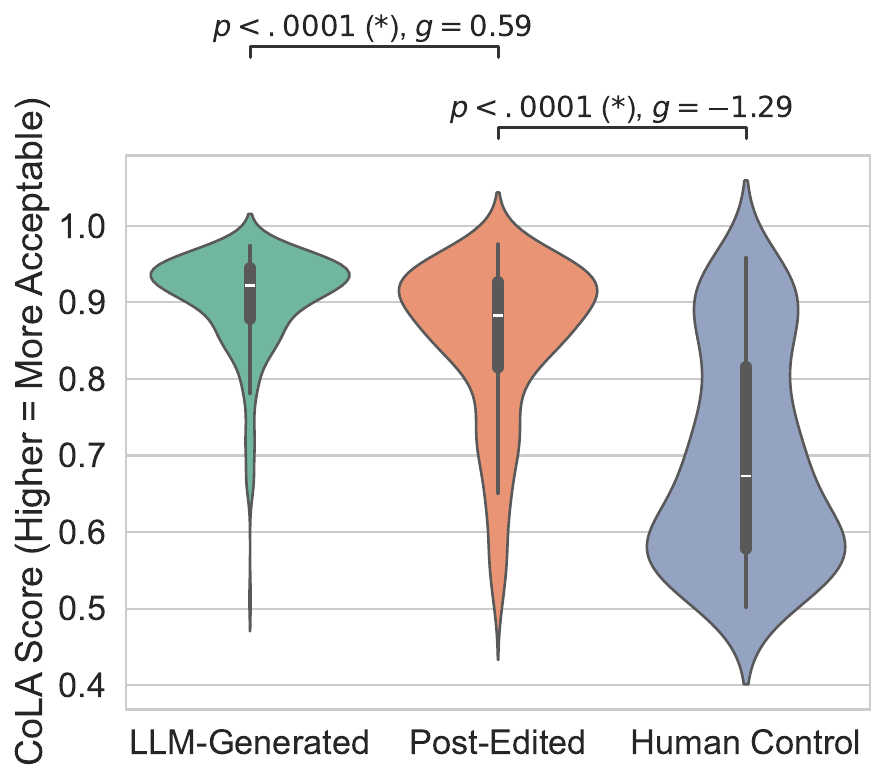}
    \caption{Linguistic acceptability (model-judged) of \modelgenerated text, post-edited text, and control text.}
    \label{fig:cola}
\end{figure}

\begin{table*}[tb]
\centering
\small
\begin{tabular}{lrrrrr}
\toprule
Predictor & Estimate & SE & z & p & 95\% CI \\
\midrule
Intercept & 0.015 & 0.004 & 3.751 & <.001 & [0.007, 0.023] \\
reference author (self) & 0.027 & 0.004 & 6.079 & <.001 & [0.018, 0.035] \\
CoLA difference & -0.086 & 0.031 & -2.777 & .005 & [-0.147, -0.025] \\
CoLA difference $\times$ reference author (self) & -0.116 & 0.040 & -2.931 & .003 & [-0.193, -0.038] \\
\bottomrule
\end{tabular}
\caption{Parameter estimates from a mixed-effects model predicting change in stylistic similarity through post-editing from reference source (same vs different participant) and change in linguistic acceptability.}
\label{tab:cola_mem}
\end{table*}

As we discussed in \autoref{sec:qual}, participants' edits often included less linguistically acceptable text including non-standard spelling, punctuation, and grammar. These features may have been unintentional (i.e., typos) or purposeful individual choices. We similarly noted that these features were often present in participants' unassisted control text. 
If these unconventional edits are similar to those in a participant's control text, we would then expect these edits to increase self-similarity. Indeed we observe many such cases.  For example, one participant adds \begin{displayquote}
...Your kindness\textcolor{red}{,e}mpathy and generosity inspire me everyday. Here's to \textcolor{red}{anotjer} year of adventures\textcolor{red}{,l}ate night calls and making memories together.
\end{displayquote}
with a typo in the word ``anotjer'' and a missing space after two commas. We observe that this participant did not put spaces after any of the six commas in their control text. 

Before the participant made this addition, the similarity of the \modelgenerated text to their control text was $.57$, and adding in these two lines increased this similarity to $.80$, a $1.4\times$ increase in self-similarity. However, after changing these potential typos, the self-similarity becomes $.70$ meaning that removing the effect of less linguistically acceptable edits leads to a more modest $1.2\times$ increase in self-similarity.

In the example above, the participant's control text accounted for six of the thirteen commas without spaces present in the full set of control text (with there being $861$ commas with spaces in the full set of control text). This may not be an error as the participant may perceive this feature as meaningful to their style, which suggests that choices in linguistic conventions can be distinct between authors and not just a generic signal of human-ness.

To exploratorily analyze the effects of less linguistically acceptable post-edits on a larger scale we collect linguistic acceptability judgments of all control, \modelgenerated, and post-edited text using a model from \citet{morris-etal-2020-textattack}\footnote{Distributed under an MIT license.} that was finetuned using expert human judgments from \citet{warstadt2019neuralnetworkacceptabilityjudgments}'s Corpus of Linguistic Acceptability (CoLA).

First, we confirm that post-edited treatment text is indeed judged as less linguistically acceptable than the original \modelgenerated text ($p<.0001$, $g=0.59$, $95\%$ CI: [$0.45$, $0.73$]) and that post-edited text is more linguistically acceptable than fully human-authored control text ($p<.0001$, $g=-1.29$, $95\%$ CI: [$-1.57$, $-1.05$]) (see \autoref{fig:cola}).

However, as we've seen, the less linguistically acceptable edits made by different participants may be qualitatively different. We fit a linear mixed effect model predicting the difference in stylistic similarity change between original \modelgenerated text and post-edited text summarized in \autoref{tab:cola_mem}. We find that changes in CoLA acceptability scores were strongly associated with changes in stylistic similarity. When similarity was measured against a \textit{different} participant's writing, a greater decrease in CoLA score (more negative CoLA difference) predicted a significantly larger increase in similarity ($\beta = -0.086$, SE $= 0.031$, $z = -2.78$, $p = .005$). This relationship was substantially stronger when similarity was measured against a participant's own control text, as indicated by a significant interaction ($\beta = -0.116$, SE $= 0.040$, $z = -2.93$, $p = .003$). In other words, post-editing text to be less linguistically acceptable increases its resemblance to human reference writing in general, but these edits make the text resemble the participant's own control writing even more than they resemble other participants' writing, perhaps due to individual-specific patterns in the kinds of less linguistically acceptable text introduced. We also note that similarity to a participant's own control text increased significantly more than similarity to others' control texts regardless of CoLA change ($\beta = 0.027$, SE $= 0.004$, $z = 6.08$, $p < .001$), consistent with our findings for H1a$^\prime$ in \autoref{sec:h1}.

\begin{figure*}[tb]
    \begin{subfigure}[t]{.48\linewidth}
\centering
    \includegraphics[height=2in, clip,trim=0 0 0 0]{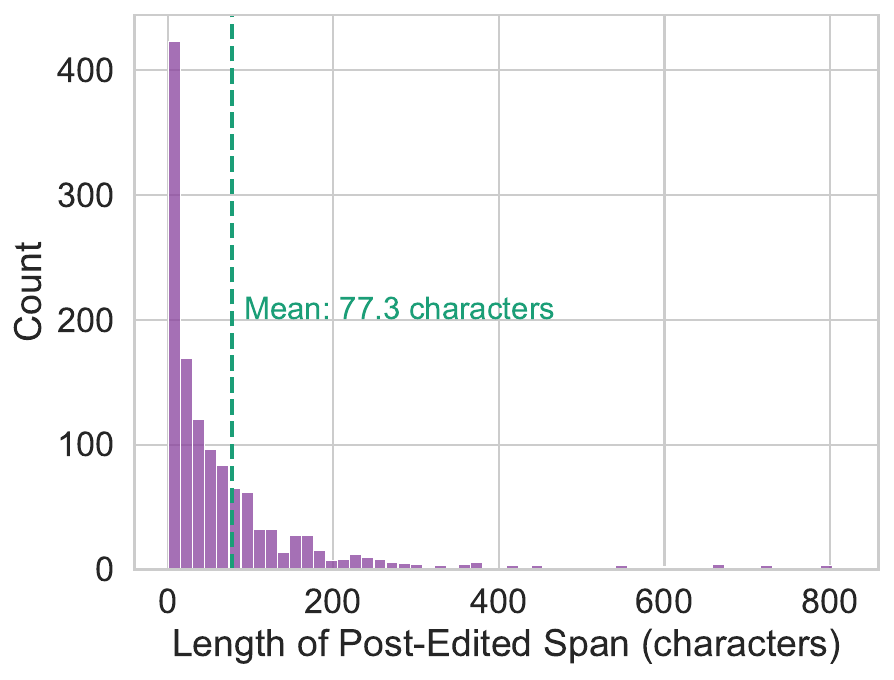}
    \caption{Distribution of post-edit span lengths}
    \Description{}
    \label{fig:edit_length_hist}    
\end{subfigure}
\hspace{.02\linewidth}
\begin{subfigure}[t]{.48\linewidth}
\centering
    \includegraphics[height=2in, clip,trim=0 0 0 0]{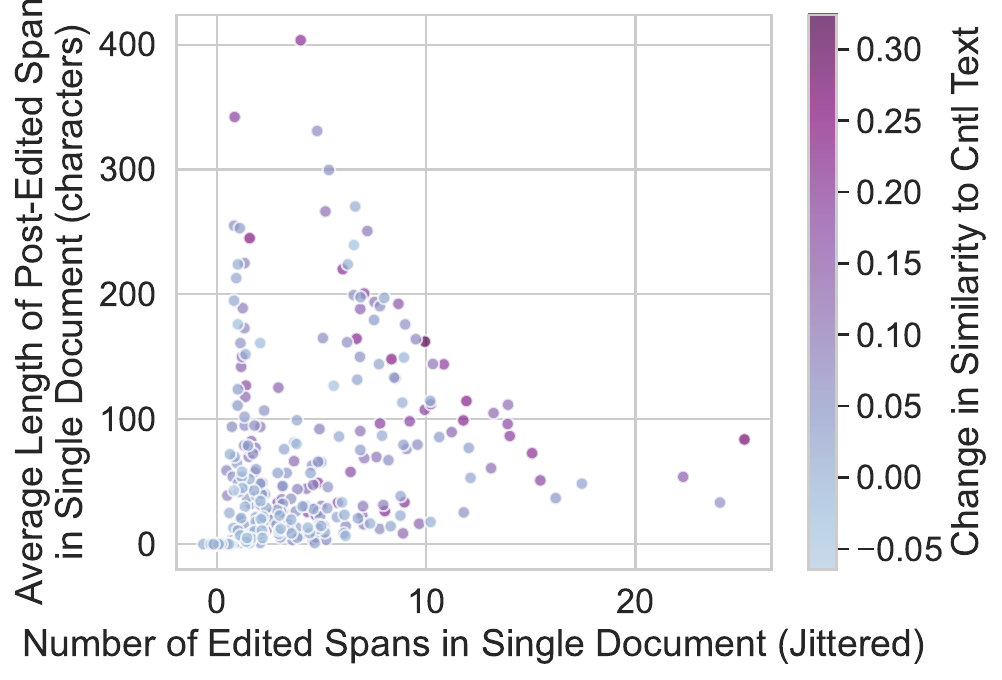}
    \caption{Document-level number vs length of post-edited spans and the resulting change in self-similarity.}
    \Description{}
    \label{fig:edit_length_scatter}  
\end{subfigure}
\caption{Length and number of post-edited spans. We see that a) spans are skewed such that the average span is somewhat long (77.3 characters) and that b) ``effective'' post-editing (deeper purple) does not clearly favor many short and diffuse edits or fewer longer edits.}
\end{figure*}

\subsubsection{Post-Editing of Lexical Features and Punctuation}\label{sec:lexical}

Here we consider the usage and post-editing of words and punctuation that have been associated with AI and human writing to understand which, if any, participants added or removed when trying to capture their personal style. We measure the frequency of LLM-associated words, em dashes (LLM-associated), and contractions (human-associated) based on reporting from the AI-detection platform Pangram\footnote{\url{https://www.pangram.com/dashboard-resources/comprehensive-guide-to-spotting-ai-writing-patterns}}.

We see that post-edited text had $5\times$ more contractions than \modelgenerated drafts suggesting that contractions may be a feature that our participants felt was important to their personal styles. 

The word ``delve'' and its variants is not present in any text written for this study. However, we do see that participants occasionally remove other words associated with \modelgenerated writing and that some of these words do indeed appear more often in \modelgenerated text than control text. Words such as ``exploring'', ``guiding'', and ``understanding'' appear about $20-50$ times in the \modelgenerated text and no more than once in the control text (though the full set of \modelgenerated text contains about $2.9\times$ as many words as the control text as we collect fewer control documents). However, in the about $15-30$ times these words appeared in \modelgenerated that participants could post-edit, they were removed a total of $2$ times each. Certain words, such as ``dive'' and ``endeavor'', that appeared fewer times in \modelgenerated text (both less than $10$) were post-edited out completely. Overall, while prior evidence shows that some of these words are statistically associated more with AI text than human text (and we see some evidence of some of these words being comparatively less frequent in the control text in our study), we see that participants do not often remove them. This may be because some of these words are not as clearly associated with AI in the popular conscious or that even if participants were aware of this association, the particular usages in our study did not feel inauthentic to their writing styles. On the other hand, we find that during post-editing, participant remove $23\%$ of the $254$ em dashes present in \modelgenerated drafts, perhaps as this difference is more well-known to the general public or feels more stylistically intrusive.

\subsubsection{Length and Diffuseness of Edits}\label{sec:edit_len}

We qualitatively observe that some participants make small, localized edits while others make longer insertions and deletions up to the paragraph level. In \autoref{fig:edit_length_hist}, we observe a long tail of edit lengths leading to a long average edit length of $77.3$ characters. In \autoref{fig:edit_length_scatter}, we consider the change in stylistic self-similarity on documents with varied number of edits and average length of edits. We exploratorily observe that both the number of edited spans in a document ($r=0.306$, $p<.0001$) and the average length of the edited spans ($r=0.207$, $p=.0011$) both positive correlate with the change in self-similarity. While our participants both used the strategy of making fewer and longer edits vs more and shorter edits, both strategies appeared to be effective. Regardless of the distribution of edits throughout the documents, we also exploratorily observe a significantly positive correlation between the overall number of changed characters and change in self-similarity ($r=0.317$, $p<.0001$). 

While one may assume that making dense edits (and especially dense additions) may not meaningfully change the style of the overall document since large swaths of text remain in the original style, these results suggest that style embedding models may be similarly sensitive to both dense and diffuse edits. This may be a limitation of these models as they are trained under the assumption that text will be written by a single author with a consistent style.

\section{Formative Survey on Writing Tasks and Importance of Personal Style}\label{sec:formative}

\begin{figure*}[tb]
    \centering
    \includegraphics[width=0.99\linewidth,  clip,trim= 5 0 5 0]{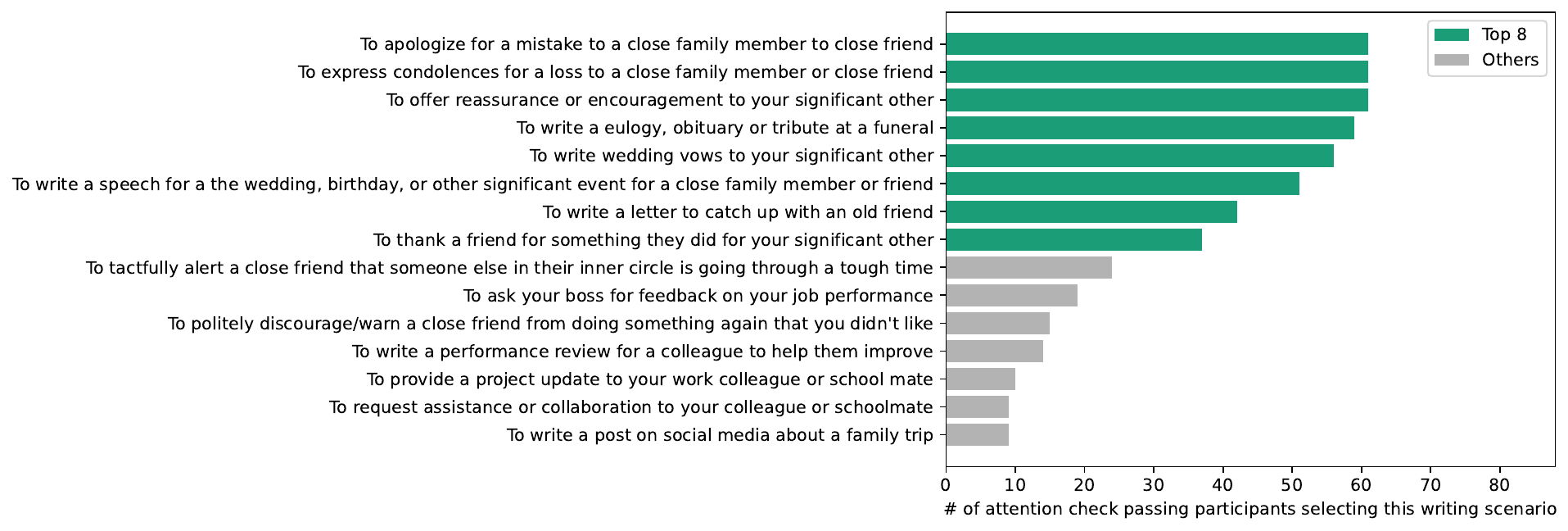}
    \caption{Number of participants selecting each candidate scenario as one where personal style is important. The ``Top $8$'' scenarios (in teal) are included in the main study.}
    \Description{}
    \label{fig:top_8}
\end{figure*}

\begin{figure*}[tb]
\begin{subfigure}[t]{.48\linewidth}
\centering
    \includegraphics[height=2in, clip,trim=0 0 0 40]{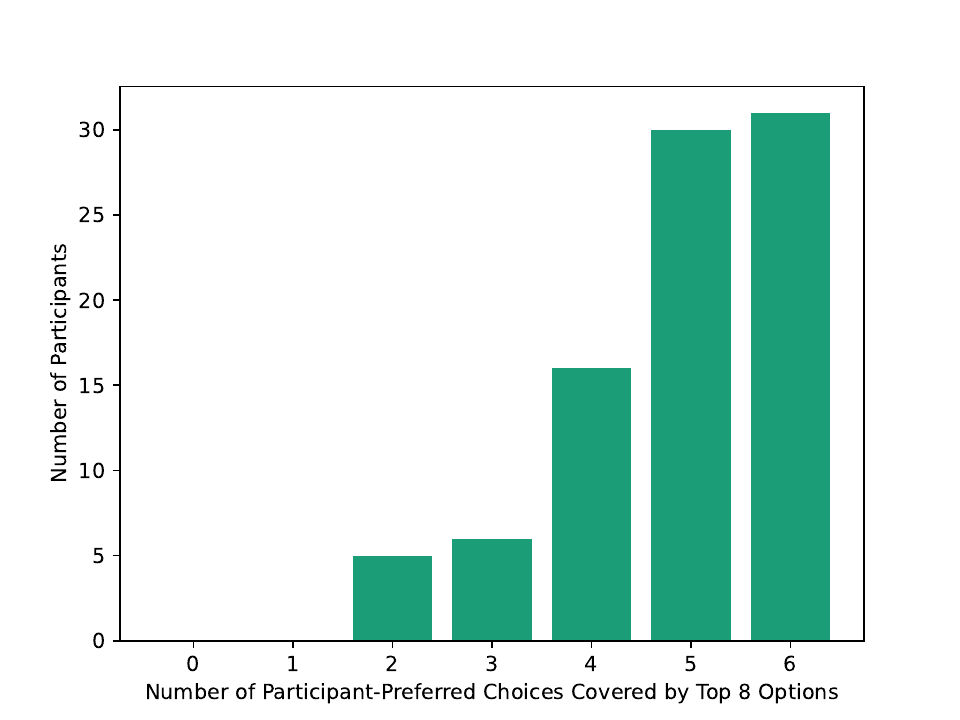}
    \caption{After selecting the top eight scenarios, how many participants have $x$ of their top choices represented? In the main study, a participant with only $x=2$ of their top choices represented would need to pick four additional writing scenarios to complete. A participant with $x=6$ of their top choices represented would not need to pick any new writing scenarios.}
    \Description{}
    \label{fig:task_coverage}  
\end{subfigure}
\hspace{.02\linewidth}
\begin{subfigure}[t]{.48\linewidth}
\centering
    \includegraphics[height=2in, clip,trim=0 0 0 5]{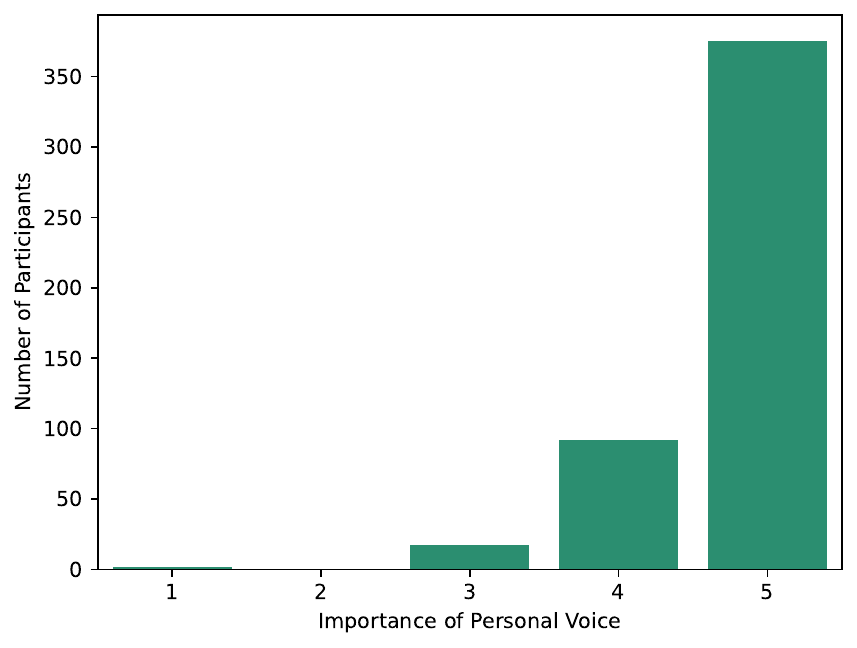}
    \caption{In practice, how important did participants in the main study say personal style is in their $6$ chosen writing scenarios? Scores are reported on a $5$-point likert scale with $1$ being very unimportant and $5$ being very important.}
    \Description{}
    \label{fig:task_importance}  
\end{subfigure}
\caption{Importance of personal style in selected tasks in a) the formative survey and b) the main study.}
\end{figure*}

Our main study included writing scenarios in which participants care about their personal style coming across. We ran a formative survey to choose appropriate writing scenarios. Participants were given 15 writing scenarios (and two attention-check scenarios) that were chosen by the authors. These tasks were shown to participants in a randomized order. Participants were asked to chose eight tasks (six actual tasks and the two attention check options) where it matters most to them that their personal style be reflected in their writing. 

$102$ participants completed this formative survey of which $88$ passed the attention checks. %
Formative study participants were also recruited from Prolific, and were not allowed to also complete the main study. The median study completion time was $3.53$ minutes, and participants were paid $\$1.25$.

We show in \autoref{fig:top_8} the number of participants who selected each writing scenario. Based on these results, we pick a set of eight scenarios to use in the main study (also shown in \autoref{tab:tasks}). 
In \autoref{fig:task_coverage}, we consider how many of the formative survey participant's chosen tasks are present in the top eight. We observe that most formative survey participants have five or six of their preferred options in the top eight (with an average of $4.86$). This means that, if they were to participate in the main study, most would either not have to select any scenarios where they feel personal style is not as important or would have to select only one. And in practice, we see in \autoref{fig:task_importance} that participants in the main study generally believe style to be very important in the writing scenarios they chose, with an average rating of $4.72\pm0.56$ on a $5$-point likert scale.

 \begin{table}[tb]
    \centering
    \footnotesize
    \begin{tabular}{p{.45\linewidth}cc}
        \toprule
        & \multicolumn{2}{c}{\# Observations} \\ 
        \cmidrule{2-3}
        Writing Task & Control & Treatment\\
        \midrule
        Thank-you letter & 25 & 53 \\
        \rowcolor{customrowcolor}Apology letter & 30 & 46 \\
        Wedding or birthday speech & 21 & 52 \\
        \rowcolor{customrowcolor}Catch-up letter & 18 & 50 \\
        Letter of condolence & 25 & 36 \\
        \rowcolor{customrowcolor}Message of reassurance or encouragement & 22 & 38 \\
        Wedding vows & 13 & 38 \\
        \rowcolor{customrowcolor}Eulogy, obituary, or tribute & 8 & 11 \\
        \bottomrule
    \end{tabular}
    \caption{Main study writing tasks. Each of the $81$ participants chose six tasks to complete. Four were randomly selected as treatment and two as control.}
    \label{tab:tasks}
\end{table}

\FloatBarrier

\section{User Study Interface}\label{sec:screenshots}

Below we include screenshots of our user study interface. We note that this interface uses the terminology such as ``your own voice'' instead of ``personal style'' as we believed this would be more understandable to lay users. 

The consent form for the study explains the study procedures as follows:

\begin{displayquote}
    In an initial survey, you will be asked to select from a list of writing tasks where writing in a way that “sounds like you” may be important.

    In the writing task, you will first be asked to provide some details about a given writing scenario. For example, for an apology letter, you might write what you did wrong. In this task, we encourage you not to include any true personal information and to instead to make up details as you see fit. \textbf{The text you write in this study may be publicly released}, so please do not include any details you may wish to remain private. After providing this planning phase, you will either be asked to complete the full writing task alone or be asked to edit an chatbot-generated copy, with the goal of making a final product that sounds authentic. Between tasks, you will be asked to answer a number of survey questions about your experience and perceptions.

    After completing a series of tasks, you will be given a post survey that includes some optional demographic questions. The study may take approximately 60 minutes to complete.

\end{displayquote}

In this consent form (as well as in the study itself), we inform participants that there writing may be publicly released and are asked not to include any true or identifying details they would like to keep private. 

After the consent form and pre-survey, participants are shown an interactive tutorial walking them through the interface. We include here the key instructions for writing details:
\begin{displayquote}
Before writing your document, we'd like you to brainstorm some of its content in the form of bullet points. Do \textbf{not} write the final full document at this stage. That step will come after planning the details. You will occasionally be given AI support in your writing, and these bullets will be used automatically to generate draft documents. 

Pasting into these bullets is not allowed in this study.
\end{displayquote}

When they are later shown an example post-editing task, they are told that ``In this example, you've been given an AI generated document to edit to sound more like you. You may instead be given an empty box to write the document from scratch.''

\begin{figure*}[ht!]
    \centering
    \includegraphics[width=.9\textwidth,clip,trim=240 0 240 0]{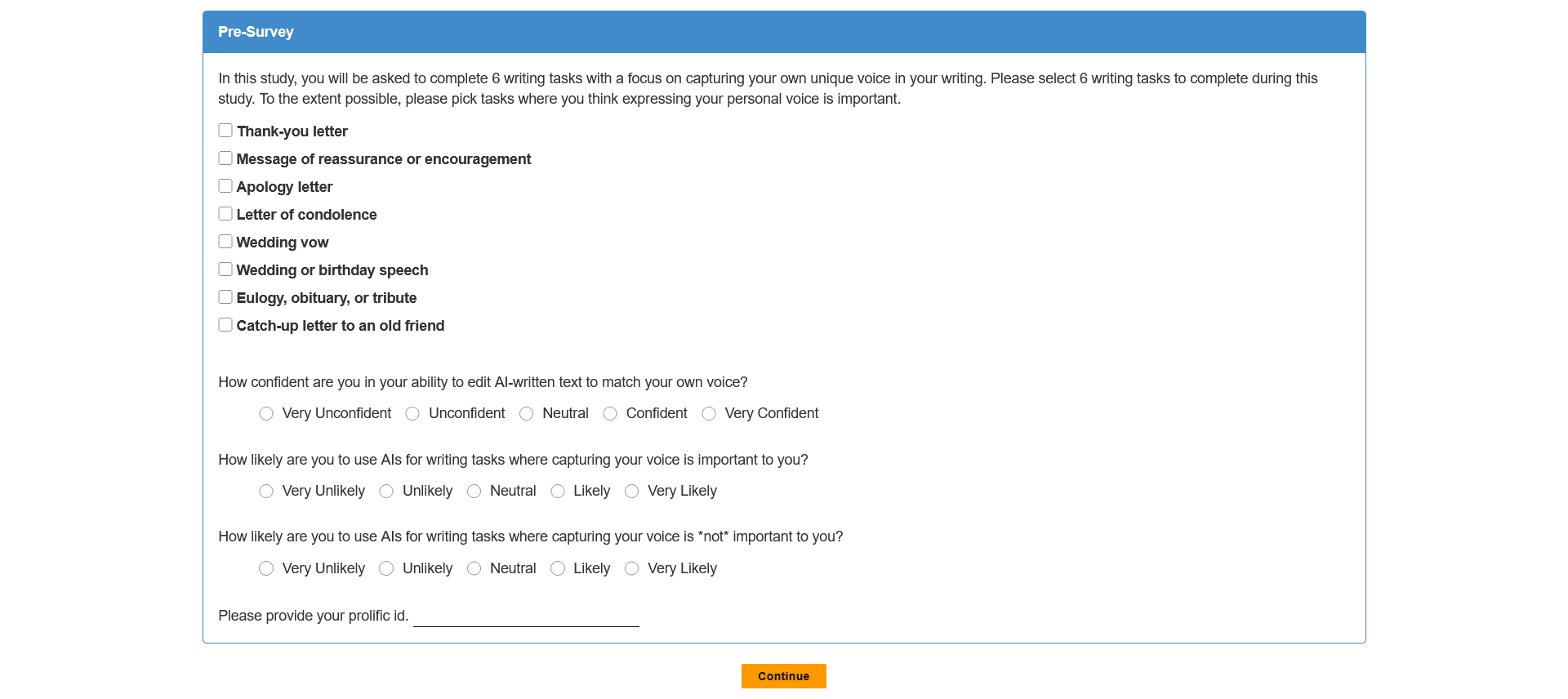}
    \caption{Pre-survey}
    \label{fig:pre_survey}
    \Description{}
\end{figure*}

\begin{figure*}[ht!]
    \centering
    \includegraphics[width=.9\textwidth,clip,trim=0 0 0 0]{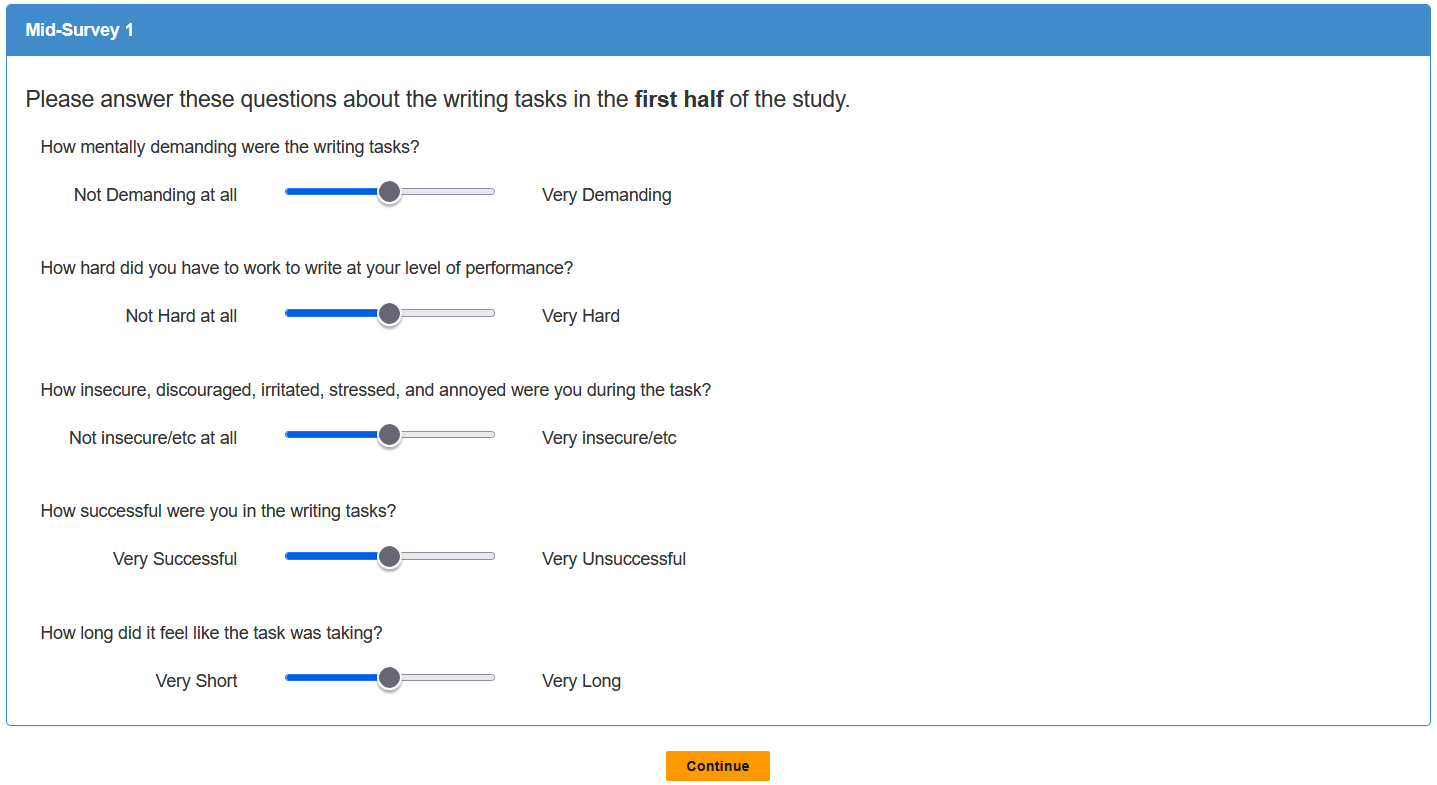}
    \caption{Task difficulty survey. An identical survey is provided after the treatment and control task blocks.}
    \label{fig:mid_survey}
    \Description{}
\end{figure*}

\begin{figure*}[ht!]
    \centering
    \includegraphics[width=.9\textwidth,clip,trim=240 0 240 1150]{interface_screenshots/post_full.png}
    \caption{Second half of the post-study survey. In the interface, both these questions and those in \autoref{fig:post_survey_1} appear on a single screen.}
    \label{fig:post_survey_2}
    \Description{}
\end{figure*}

\begin{figure*}[ht!]
    \centering
    \begin{subfigure}[c]{\textwidth}
        \centering
        \includegraphics[width=.9\textwidth,clip,trim=240 0 240 0]{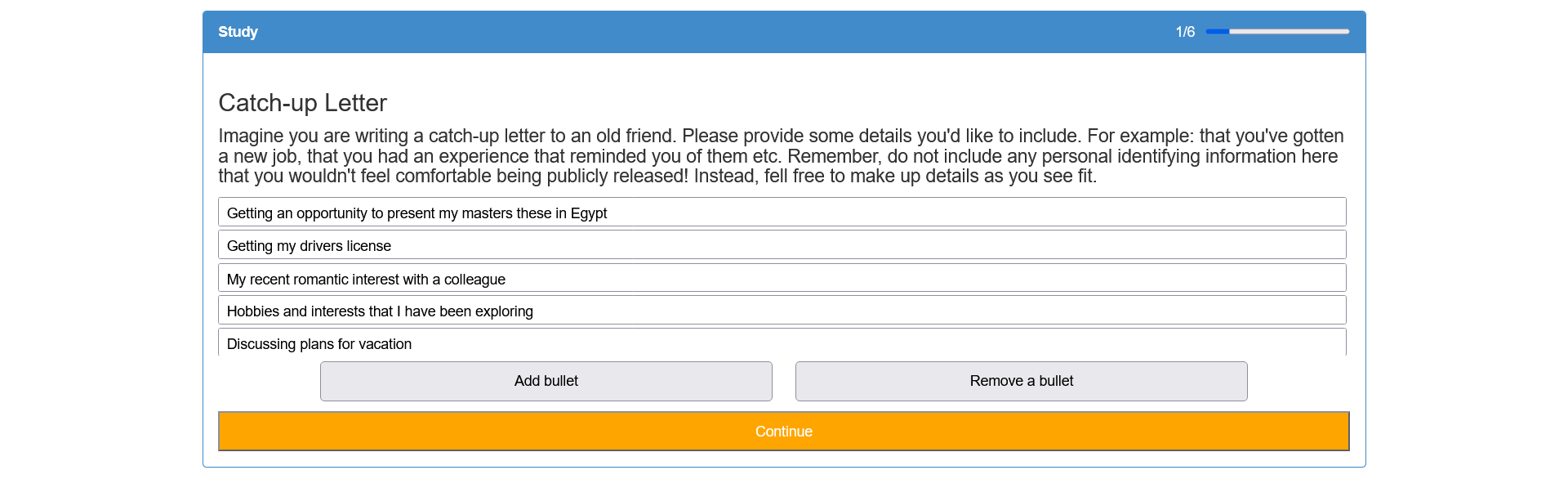}
        \caption{}
    \end{subfigure}
    \begin{subfigure}[c]{\textwidth}
        \centering
        \includegraphics[width=.9\textwidth,clip,trim=240 0 240 0]{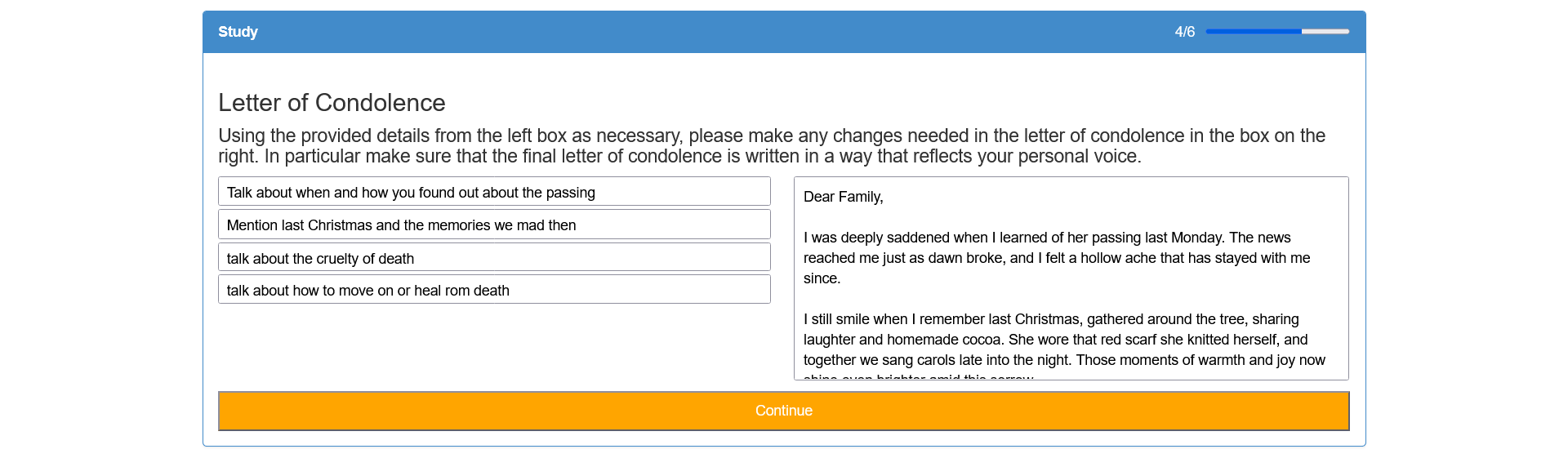}
        \caption{}
    \end{subfigure}
    \caption{Sample study task. In a) the participant has finished planning the details to include in their catch-up letter, and in b) they have finished post-editing the \modelgenerated condolence letter. Unless or until the participant asks not to be reminded anymore, a popup reminds them to make sure they are not including any true details when they try to submit details.}
    \label{fig:task}
    \Description{}
\end{figure*}

\begin{figure*}[ht!]
    \centering
    \includegraphics[width=.9\textwidth,clip,trim=240 0 240 0]{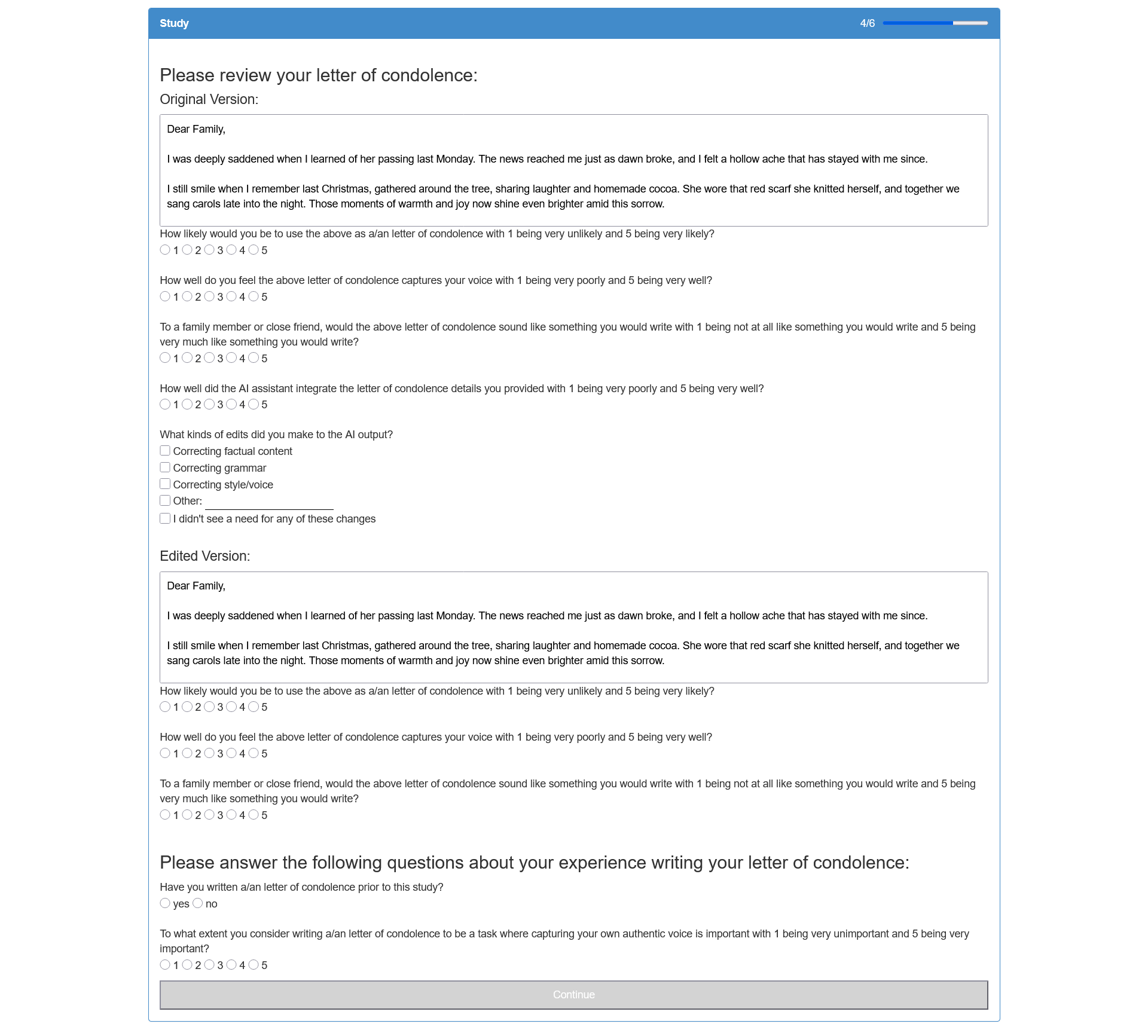}
    \caption{Post-task survey questions. In the control block, participants are not asked about the \modelgenerated ``Original Version''.}
    \label{fig:survey_edit}
    \Description{}
\end{figure*}

\FloatBarrier

\end{document}